\documentclass{article}
\usepackage[nonatbib, final]{nips_style}
\usepackage[utf8]{inputenc} % allow utf-8 input
\usepackage[T1]{fontenc}    % use 8-bit T1 fonts
\usepackage{hyperref}       % hyperlinks
\usepackage{url}            % simple URL typesetting
\usepackage{booktabs}       % professional-quality tables
\usepackage{amsfonts}       % blackboard math symbols
\usepackage{nicefrac}       % compact symbols for 1/2, etc.
\usepackage{microtype}      % microtypography
\usepackage{float}
\usepackage{subcaption}
\usepackage{graphicx}
\usepackage{cite}
\usepackage{enumitem}
\usepackage[table,xcdraw]{xcolor}
\usepackage{adjustbox}
\usepackage{hyperref}
\usepackage{svg}
\usepackage{sectsty}
\usepackage{titlesec}
\usepackage{url}
\usepackage{float}
\usepackage{microtype}
\usepackage{graphicx} % Required for inserting images
\usepackage{amsmath}
\usepackage[utf8]{inputenc} % allow utf-8 input
\usepackage[T1]{fontenc}    % use 8-bit T1 fonts
\usepackage{hyperref}       % hyperlinks
\usepackage{url}            % simple URL typesetting
\usepackage{booktabs}       % professional-quality tables
\usepackage{amsfonts}       % blackboard math symbols
\usepackage{nicefrac}       % compact symbols for 1/2, etc.
\usepackage{microtype}      % microtypography
\usepackage{xcolor}         % colors
\usepackage{xspace}
\usepackage{amsmath}
\usepackage{amssymb}
\usepackage{booktabs}
\usepackage{makecell}
\usepackage{colortbl}
\usepackage{graphicx}   
\usepackage{CJKutf8}
\usepackage{multirow}
\usepackage{interval}
\intervalconfig{soft open fences}
\usepackage{algorithm, algpseudocode}

\usepackage{float}
\usepackage{enumitem}
\usepackage{tabularx}
\usepackage{placeins}
\usepackage{color}
\usepackage{tcolorbox}
\usepackage{verbatim}
\usepackage{colortbl} 
\usepackage{listings}
\usepackage{makecell}
\usepackage{siunitx}
\usepackage{xcolor}
\definecolor{bg}{RGB}{176,226,255}
\definecolor{bonus_green}{RGB}{0,100,0}
\newcommand{\bbonus}[1]{{\textcolor{bonus_green}{$^{\uparrow#1}$}}}

\newcommand*\samethanks[1][\value{footnote}]{\footnotemark[#1]}

\title{Logic-RL: Unleashing LLM Reasoning with Rule-Based Reinforcement Learning}
\author{
Tian Xie$^{1}$\thanks{~~Work done during internship at MSRA. Open-Source Research Project.} \quad
Zitian Gao$^{2}$ \quad
\textbf{Qingnan Ren}$^{3}$ \quad
Haoming Luo$^{3}$ \quad
\textbf{Yuqian Hong}$^{1}$\samethanks\\
\textbf{Bryan Dai}$^{2}$ \quad
\textbf{Joey Zhou}$^{2}$ \quad
\textbf{Kai Qiu}$^{1}$ \quad
\textbf{Zhirong Wu}$^{1}$ \quad
\textbf{Chong Luo}$^{1}$\thanks{~~Corresponding author.} \vspace{2mm}\\
$^1$Microsoft Research Asia \quad
$^2$Ubiquant \quad
$^3$Independent \vspace{2mm} \\
\texttt{\{unakar666, hmluo65536, hoknight0\}@gmail.com} \\
\texttt{\{v-yuqianhong, kai.qiu, wu.zhirong, chong.luo\}@microsoft.com} \\
\texttt{\{ztgao02, cbdai, jzhou\}@ubiquant.com} \\
}

\begin{document}
\maketitle

% %%%% authorstart
% \author{
% Tian Xie$^{1,4}$\thanks{~~Work done during internship at MSRA.} 
% \quad Zitian Gao$^{2}$ 
% \quad \textbf{Qingnan Ren}$^{3}$
% \quad Haoming Luo$^{3}$ 
% \quad \textbf{Yuqian Hong}$^{1,4}$\\ 

% \textbf{Bryan Dai}$^{2}$
% \quad \textbf{Jay Zhou}$^{2}$ 
% \quad \textbf{Kai Qiu}$^{1}$  
% \quad \textbf{Chong Luo}$^{1}$\thanks{~~Corresponding author.} \\

\vspace{-3mm}

\begin{abstract}

Inspired by the success of DeepSeek-R1, we explore the potential of rule-based reinforcement learning (RL) in large reasoning models. 
To analyze reasoning dynamics, we use synthetic logic puzzles as training data due to their controllable complexity and straightforward answer verification.
We make some key technical contributions that lead to effective and stable RL training: a system prompt that emphasizes the thinking and answering process, a stringent format reward function that penalizes outputs for taking shortcuts, and a straightforward training recipe that achieves stable convergence.
% and a tempering-annealing strategy that balances exploration and exploitation.
Our 7B model develops advanced reasoning skills—such as reflection, verification, and summarization—that are absent from the logic corpus. Remarkably, after training on just 5K logic problems, it demonstrates generalization abilities to the challenging math benchmarks AIME and AMC.

\begin{figure}[H]
\centering
% \begin{minipage}{0.8\linewidth}
    \centering
    \hspace{-2mm}{
    \includegraphics[width=1\linewidth]{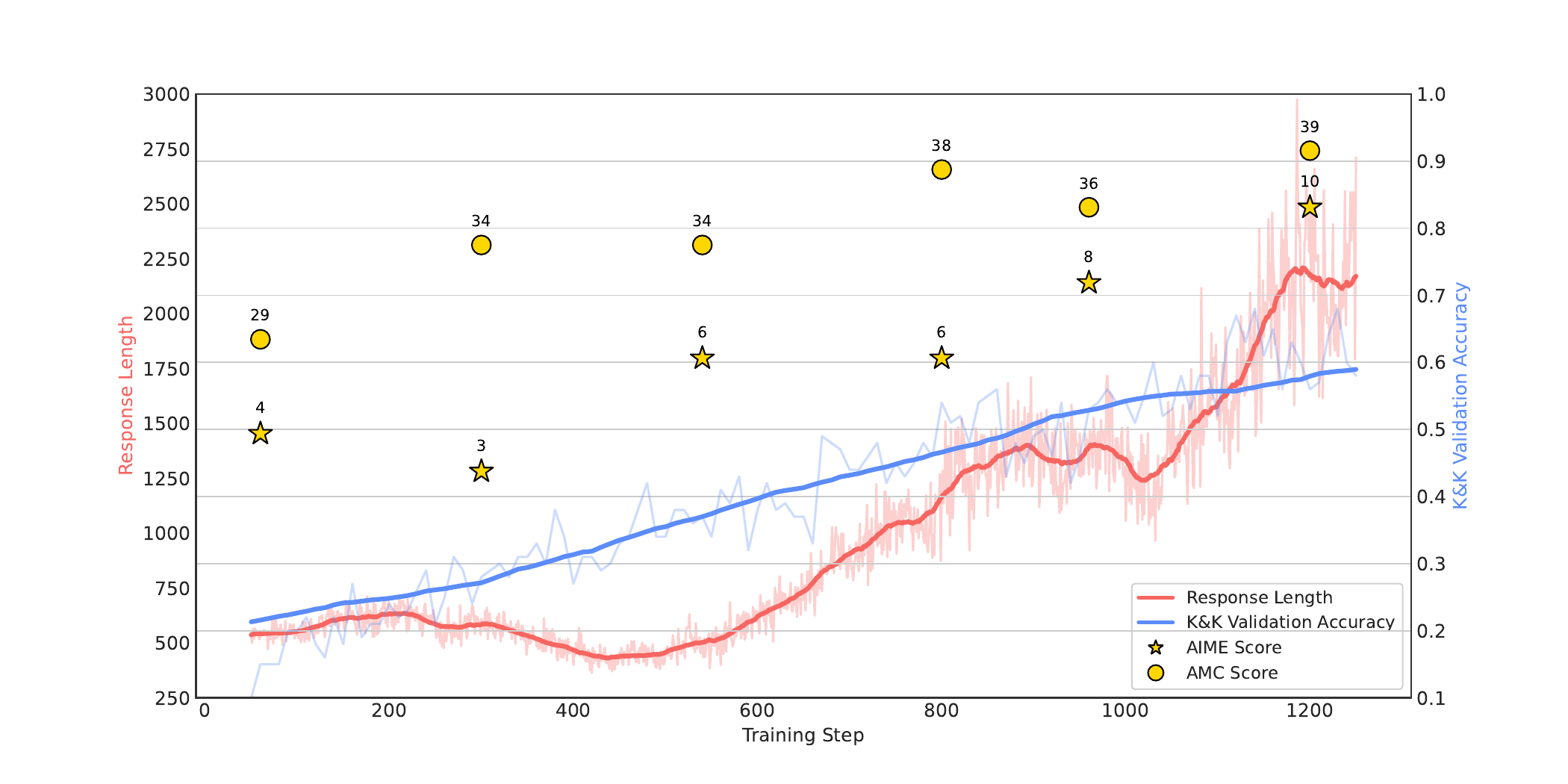}}
\caption{Validation accuracy and mean response length during RL training, illustrating how the model autonomously learns to allocate more thinking compute for improved performance. Remarkably, the model also demonstrates impressive generalization on completely unseen datasets (AIME, AMC).}
    \label{fig:len}
% \end{minipage}
\end{figure}
\end{abstract}

\section{Introduction}

The post-training phase of large language models (LLMs) has advanced rapidly~\cite{xu2025largereasoningmodelssurvey}, with models like DeepSeek-R1~\cite{deepseekai2025deepseekr1incentivizingreasoningcapability}, Kimi-K1.5~\cite{kimiteam2025kimik15scalingreinforcement}, and OpenAI-o1~\cite{o1} demonstrating remarkable reasoning abilities. DeepSeek-R1, in particular, introduced a simple yet effective rule-based reinforcement learning (RL) approach, enabling emergent reasoning patterns without relying on traditional scaffolding techniques such as Monte Carlo Tree Search (MCTS)~\cite{sc-mcts, deepprover1.5, xu2023traingainunleashmathematical, feng2024alphazeroliketreesearchguidelarge} or Process Reward Models (PRM)\cite{lightman2023letsverifystepstep}.

Despite these advancements, substantial gaps remain in translating these developments into reproducible research. While DeepSeek-R1 provides open-source model weights, it does not release the corresponding training code or dataset. This absence raises critical questions: (1) Can similar reasoning abilities emerge in smaller-scale models? (2) What is the optimal training data structure for fostering such capabilities? (3) What methodologies can reliably replicate these results?

Addressing these questions requires controlled experimental frameworks that isolate key variables.
%In this paper, we explore to unleash the reasoning capcablilites of LLMs using rule-based reinforcement learning. 
While mathematics is often treated as the common testbed for reasoning, widely-used math datasets like GSM8K\cite{cobbe2021trainingverifierssolvemath} and Omini-MATH \cite{gao2024omnimathuniversalolympiadlevel} suffer as the training data due to its uncontrolled variance in problem complexity, which may span various logical induction depths. 
To overcome this limitation, we leverage a procedurally generated Knights and Knaves (K\&K) logic puzzle dataset~\cite{memllm}, which allows  controllable difficulty levels and ease of rule-based reward verification,  making it ideal for studying reasoning dynamics.

In this paper, we introduce Logic-RL, a rule-based reinforcement learning framework that acquires
R1-like reasoning patterns through training on logic puzzles.
The training framework adopts the REINFORCE++ algorithm~\cite{rpp} and the reward designs from DeepSeek-R1 for post-training.
While naive training would lead to collapsed solutions, we propose a practical system prompt and a stringent format reward to avoid the reasoning model for taking shortcuts.
We also incorporate a few modifications to the REINFORCE++ algorithm for improved performance.

As the RL training undergoes, we observe that the model naturally allocates more training steps to reason. This computational expansion scales from generating hundreds to thousands of tokens, enabling deeper exploration and refinement of its thought processes.
We evaluate the model performance on the challenging math benchmarks for reasoning.
%Controlled experiments reveal surprising generalization: 
With merely 5,000 procedurally generated logic puzzles, our 7B model improves by 125\% on AIME and 38\% on AMC against the base model.
%, both known as hard math benchmarks. 
This cross-domain generalization capability suggests that RL-trained reasoning heuristics develop abstract problem-solving schemata rather than relying on domain-specific pattern matching.

Besides the technical contributions mentioned above, our study also makes several interesting findings:
\vspace{-5mm}
\begin{itemize}[leftmargin=*]
    \item \textbf{Longer responses don’t guarantee better reasoning.} 
    Length alone is not a valid performance metric for training time evaluation. The most efficient reasoning comes from the shortest path.
    \item \textbf{Language mixing hinders reasoning.} This observation underscores the need for a language consistency penalty in reward modeling.
    \item \textbf{Increasing `thinking' tokens do help.} RL training naturally boosts the frequency of reflection-related words, suggesting a correlation between certain tokens' frequency and performance.
    \item \textbf{SFT memorizes; RL generalizes.} SFT relies heavily on memorization, often leading to superficial shortcut learning, whereas RL self-evolves with minimal dependence on dataset structure. 
    \item \textbf{Cold start is a bonus, not a necessity.} Training dynamics remain surprisingly similar whether starting from a base or instruct model, though the latter exhibits slightly better performance.
    \item  \textbf{Curriculum Learning still matters.} Under a fixed data curation ratio, a well-designed curriculum learning approach always outperforms random shuffle.
\end{itemize}
 % \item \textbf{A simple and reproducible training framework.} A strict rule-based reward model ensures an unhackable environment, while a temper-anneal sampling strategy effectively balances exploration and exploitation.
%\item \textbf{A simple and reproducible training framework.} A strict rule-based reward model ensures an unhackable environment, while a straightforward training strategy achieves stable convergence through continuous training without complex scheduling.
\section{Method}
\subsection{Data Synthesis}

The Knights and Knaves (K\&K) puzzles \cite{memllm} constitute an algorithmically generated reasoning dataset. In these puzzles, characters are either knights, who always tell the truth, or knaves, who always lie. The objective is to determine the nature of each character based on their statements. This dataset is distinguished by its high degree of controllability:

\begin{enumerate}[leftmargin=*]
    \item \textbf{Procedural Generation}: Puzzles are systematically generated using logic templates, ensuring both consistency and infinite variability. Importantly, these puzzles represent unseen data for the original model, making them ideal for testing generalization capabilities.
    
    \item \textbf{Controlled Difficulty Levels}: The difficulty of the puzzles can be precisely adjusted, enabling the design of a curriculum learning strategy. Difficulty is modulated by varying the number of characters (2–8) and the complexity of logical operations (1–4 combinations of Boolean operators). Furthermore, more complex puzzles can serve as out-of-distribution tests for models trained on simpler cases, providing insights into their ability to generalize.

    \item \textbf{Ease of Verification}: Each puzzle has a single, unambiguous ground truth answer, with correctness guaranteed by the generation algorithm. Solutions require strict deductive reasoning, allowing for accurate evaluation of model responses and minimizing the risk of reward hacking.
\end{enumerate}

% example就换成橙色吧，跟system prompt区分一下
% \begin{tcolorbox}[
%     colframe=orange!50!white,
%     colback=orange!5,
%     coltitle=black,
%     fonttitle=\bfseries,
%     title=An example of a K\&K puzzle\label{long_open_q},
% ]

\begin{tcolorbox}[
    % enhanced, % 启用增强模式以支持更多功能
    colframe=blue!70!black, % 边框颜色：深蓝色带一点黑色
    colback=blue!10!white, % 背景颜色：浅蓝色
    coltitle=white, % 标题文字颜色：白色
    fonttitle=\bfseries, % 标题字体加粗
    title=An example of a K\&K puzzle\label{long_open_q}, % 标题内容
    % attach boxed title to top left={ % 将标题放置在左上角
    %     yshift=-2mm, % 垂直偏移
    %     xshift=2mm % 水平偏移
    % },
    % boxed title style={ % 标题框样式
    %     colframe=blue!70!black, % 标题框边框颜色
    %     colback=blue!70!black % 标题框背景颜色
    % },
    sharp corners, % 圆角设置为锐角
    boxrule=0.5mm, % 边框宽度
    % drop fuzzy shadow=gray!30 % 添加柔和阴影
]
\textbf{Problem}:  
A very special island is inhabited only by knights and knaves. Knights always tell the truth, and knaves always lie. You meet 2 inhabitants: Zoey, and Oliver. Zoey remarked, "Oliver is not a knight". Oliver stated, "Oliver is a knight if and only if Zoey is a knave". So who is a knight and who is a knave? \\

\textbf{Solution}:  
(1) Zoey is a knave (2) Oliver is a knight
\label{puzzle}
\end{tcolorbox}

\vspace{2mm}
The K\&K puzzle \ref{puzzle} is exceptionally well-suited for further analysis due to its synthetic design and logical precision. Each puzzle is constructed using formal rules, ensuring that every problem has a unique solution that can be deterministically verified. This eliminates the ambiguities commonly encountered in natural language tasks, enabling us to clearly distinguish between genuine reasoning capabilities and superficial memorization.

\subsection{Rule Based Reward Modeling}
The reward serves as the primary training signal in reinforcement learning (RL), guiding the optimization process. We continuously monitored hacking behaviors in the model's outputs, refining our reward design iteratively. This led to a nearly unhackable, rule-based reward system that comprises only two types of rewards: Format Reward and Answer Reward.
\begin{tcolorbox}[
    colframe=teal!70!black, % 边框颜色：深青色带一点黑色
    colback=teal!10!white, % 背景颜色：浅青色
    coltitle=white, % 标题文字颜色：白色
    fonttitle=\bfseries, % 标题字体加粗
    title=System Prompt\label{long_open_q}, % 标题内容
    % attach boxed title to top left={ % 将标题放置在左上角
    %     yshift=-2mm, % 垂直偏移
    %     xshift=2mm % 水平偏移
    % },
    % boxed title style={ % 标题框样式
    %     colframe=teal!70!black, % 标题框边框颜色
    %     colback=teal!70!black % 标题框背景颜色
    % },
    sharp corners, % 圆角设置为锐角
    boxrule=0.5mm, % 边框宽度
    % drop fuzzy shadow=gray!30 % 添加柔和阴影
]
You are a helpful assistant. The assistant first thinks about the reasoning process in the mind and then provides the user with the answer. The reasoning process and answer are enclosed within <think> </think> and<answer> </answer> tags, respectively, i.e., <think> reasoning process here </think><answer> answer here </answer>.  Now the user asks you to solve a logical reasoning problem. After thinking, when you finally reach a conclusion, clearly state the identity of each character within <answer> </answer> tags. i.e., <answer> (1) Zoey is a knight, (2) ... </answer>.
\label{prompt}
\end{tcolorbox}

\paragraph{Format Reward:} We use regular expression extraction to enforce a structured response format. The model is required to put its reasoning process within \texttt{<think></think>} tags and provide the final conclusion inside \texttt{<answer></answer>} tags. Additionally, we recommend including a \texttt{<think>} tag directly at the end of the prompt, which significantly reduces the difficulty for the base model to follow our instructions.

% To ensure strict adherence to this structure, we enforce that each tag appears exactly once and in the correct sequential order.  This design effectively prevents potential reward hacking behaviors, such as:

Under our early imperfect  rule design, we consistently observed  reward hacking phenomena, some of which are listed below:

\begin{itemize}[leftmargin=*]
    \item Skipping the \texttt{<think></think>} process and directly answering.
    \item Placing reasoning inside the \texttt{<answer></answer>} tag.
    \item Repeatedly guessing answers without proper reasoning.
    \item Including irrelevant nonsense in addition to providing the answer.
    \item Organizing correct answer in a wrong manner for extraction.
    \item Revisiting the thinking phase after already outputting an \texttt{<answer>} due to insufficient reasoning.
    \item Repeating the original question or using phrases like "thinking process here" to avoid true reasoning.
\end{itemize}

Accordingly, we iteratively refine our rule design. For example, each tag should appear exactly once and in the correct sequential order, the thinking process must include genuine reasoning, and the conclusion should be presented in an extractable and readable manner. By enforcing these constraints, we ensure that different actions receive appropriate rewards based on their adherence to the format. The format score (\( S_{format} \)) is computed as follows:

\[
S_{format} =
\begin{cases}
\text{1}, & \text{if format is correct} \\
\text{-1}, & \text{if format is incorrect}
\end{cases}
\]

\paragraph{Answer Reward:} The second component evaluates the correctness of the content in the model’s response. Once the format is validated, we check if the model's answer matches the ground truth. The answer score (\( S_{answer} \)) is computed as:

\[
S_{answer} =
\begin{cases}
2, & \text{if the answer fully matches the ground truth} \\
-1.5, & \text{if the answer partially mismatches the ground truth} \\
-2, & \text{if the answer cannot be parsed or is missing}
\end{cases}
\]

\subsection{RL Algorithm}
We adopt a modified version of \texttt{REINFORCE++} as our baseline algorithm, which has demonstrated superior performance compared to GRPO in our experimental setup. A detailed comparison of these algorithms is provided in Section~\ref{rl_algos}. 

\textbf{Reinforce Return Calculation:} The discounted cumulative rewards for each trajectory are computed as below, where $\gamma$ is the discount factor, set to 1 in our experiments:

\begin{equation*}
G_t = \sum_{k=t+1}^{T} \gamma^{k-t} r_k
\end{equation*}

Following recommendations from DeepSeek-Math~\cite{grpo}, we incorporate several minor refinements into the implementation of \texttt{REINFORCE++}.

\paragraph{First modification: Use KL Loss}
 The KL-divergence between the response distributions of the RL model and the SFT model is calculated for each token. This divergence is incorporated as a penalty term in the reward function of PPO during training. The per-token reward is defined as follows:
\begin{equation*}
r(s_t, a_t) = \mathbf{I}(s_t = [\text{EOS}]) r(x, y) - \beta \text{KL}(t),
\end{equation*}
where $\mathbf{I}(s_t = [\text{EOS}])$ is an identity function that evaluates to 1 when the \texttt{<eos>} token is reached, and $\beta$ controls the weight of the KL penalty.

In contrast, the GRPO implementation does not include the KL-divergence as part of the reward function. Instead, it directly incorporates the KL-divergence into the loss function, arguing that this approach simplifies the computation and avoids unnecessary complexity. Following this rationale, we also use KL loss like GRPO:
\begin{align*}
    \mathcal{J}_{\text{GRPO}}(\theta) &= \mathbb{E}_{[q \sim P(Q), \{o_i\}_{i=1}^G \sim \pi_{\theta_{\text{old}}}(O|q)]} \notag \\
    &\frac{1}{G}\sum_{i=1}^G\frac{1}{\vert o_i\vert}\sum_{t=1}^{\vert o_i\vert}\Bigg\{\min\left[\frac{\pi_{\theta}^{i,t}}{\pi_{\theta_{\text{old}}}^{i,t}}\hat{A}_{i,t}, \textrm{clip}\left(\frac{\pi_{\theta}^{i,t}}{\pi_{\theta_{\text{old}}}^{i,t}}, 1-\epsilon, 1+\epsilon\right)\hat{A}_{i,t}\right] - \beta\mathbb{D}_{\text{KL}}[\pi_{\theta} \| \pi_{\text{ref}}]\Bigg\}.
\end{align*}

\paragraph{Second Modification: KL Estimation}

Another key distinction lies in how the KL-divergence is estimated. The default KL estimator for PPO is defined as follows:

\begin{equation*}
\text{KL}(t) = \log\left(\frac{\pi_{\theta_{\text{old}}}(a_t | s_t)}{\pi_{\theta}(a_t | s_t)}\right).
\end{equation*}

In contrast, GRPO employs an unbiased estimator for the KL-divergence, formulated as:

\begin{equation*}
\mathbb{D}_{\text{KL}}\left[\pi_{\theta} \| \pi_{\text{ref}}\right] = \frac{\pi_{\text{ref}}\left(o_{i, t} | q, o_{i,<t}\right)}{\pi_{\theta}\left(o_{i, t} | q, o_{i,<t}\right)} - \log \frac{\pi_{\text{ref}}\left(o_{i, t} | q, o_{i,<t}\right)}{\pi_{\theta}\left(o_{i, t} | q, o_{i,<t}\right)} - 1.
\end{equation*}

This approach ensures that the KL estimate is always non-negative, whereas the original formulation may yield negative values. GRPO's estimator provides a more stable and reliable measure of divergence during training.

After implementing these modifications, we proceed with our experiments.

\subsection{Training Schedule}
We directly train the model for 3600 steps with a constant learning rate of $4 \times 10^{-7}$ and temperature parameter of 0.7. During training, the model is directly exposed to mixed complexity logic puzzles ranging from 3 to 7 people. This straightforward training regimen achieves competitive performance as shown in the final results table. Through continuous training with these fixed hyperparameters, the model develops stable reasoning patterns characterized by logical exploration, intermediate verification, and systematic summarization before producing final answers. These emergent behaviors demonstrate the model's capacity to handle complex logical reasoning tasks effectively.

% We employ a multi-stage curriculum learning approach, gradually exposing the model to logic puzzles of increasing complexity, ranging from 3-person to 7-person problems. Both stages balance exploration and exploitation, fostering a model that generalizes effectively across diverse problems.

% \textbf{Tempering Stage:}  
% In the initial phase, we set a learning rate of $5 \times 10^{-7}$ and a temperature of $1.0$ to encourage exploration. This stage prioritizes token diversity over validation accuracy, enabling the model to explore a wide range of reasoning strategies. By breaking down the rigid markdown format used by the original model, we prevent it from converging on narrow solutions. To ensure stable updates during this exploratory phase, we use a relatively large rollout.

% \textbf{Annealing Stage:}  
% As training progresses, we gradually reduce the learning rate (first to $2 \times 10^{-7}$, then to $1 \times 10^{-7}$) and lower the temperature (first to $0.7$, then to $0.4$). This shift refines the model's policy, emphasizing accuracy and consistency in its solutions. During this phase, the model begins to exhibit stable reasoning patterns, characterized by behaviors such as exploration, hesitation, verification, and summarization before producing a final answer. These behaviors reflect a deeper understanding of problem structures and lead to more reliable outputs.

Other key parameters used during training are summarized in Table \ref{tab:training_parameters}

\begin{table}[h!]
    \centering
    \caption{Important Training Parameters}
    \label{tab:training_parameters}
    \begin{tabular}{ccccc}
        \toprule
        \textbf{Algorithm} & \textbf{Train Batch Size} & \textbf{Rollout N} & \textbf{KL Coef} & \textbf{Max Response Len} \\
        \midrule
        REINFORCE++ & 8 & 8 & 0.001 & 4096 \\
        \bottomrule
    \end{tabular}
\end{table}
% 这个table还是去掉吧，感觉两阶段文字描述就可以了
\section{Experiment}
\label{main}
\vspace{-3mm}
We began by experimenting with various models from the Qwen2.5 series as potential baseline candidates. For instance, \texttt{Qwen2.5-Math-7B} exhibited a strong tendency to generate Python code blocks, which often conflicted with our strict formatting requirements. Despite efforts to mitigate this behavior by removing system prompts and penalizing specific markdown styles, it remained challenging to fully address.

Additionally, we tested both \texttt{Qwen2.5-7B-Base} and \texttt{Qwen2.5-7B-Instruct} as starting points. Surprisingly, we found that the base and instruct models displayed nearly identical training metrics during RL training, including validation accuracy, response length growth curves, and reward curves. A detailed comparison between base \& instruct model can be found in the appendix \ref{base_comp}. However, the instruct model demonstrated slightly higher test accuracy, making it the preferred choice. Consequently, we selected \textbf{Qwen2.5-7B-Instruct-1M}~\cite{qwen251m} as our baseline. See more base \& instruct model training dynamics comparision in Appendix \ref{fig:base_length}

Notably, despite the training dataset being limited to 3 to 7-person K\&K logic puzzles—with fewer than 5,000 synthetic samples—the model demonstrates a remarkable ability to generalize to out-of-distribution (OOD) scenarios, such as 8-person puzzles.

\begin{table}[h]
    \centering
    \renewcommand{\arraystretch}{1.3}
    \Large
    \resizebox{\linewidth}{!}{
    \begin{tabular}{@{}l *{8}{c} @{}}
    \toprule
    \multirow{3}{*}{\centering\textbf{Model}} & \multicolumn{6}{c}{\textbf{Difficulty by Number of People}} \\
    \cmidrule(r){2-8}
     & \multicolumn{1}{c}{2} & \multicolumn{1}{c}{3} & \multicolumn{1}{c}{4} & \multicolumn{1}{c}{5} & \multicolumn{1}{c}{6} & \multicolumn{1}{c}{7} & \multicolumn{1}{c}{8} & \multicolumn{1}{c}{\multirow{2}{*}[+2.5ex]{Avg.}} \\
    \midrule

    \textbf{o3-mini-high} & 0.99 & 0.98 & 0.97 & 0.95 & 0.94 & 0.89 & 0.83 & 
 0.94 \\
    \textbf{o1-2024-12-17} & 0.83 & 0.51 & 0.38 & 0.38 & 0.35 & 0.30 & 0.20 & 0.42\\
    \textbf{Deepseek-R1} & 0.91 & 0.73 & 0.77 & 0.78 & 0.75 & 0.88 & 0.83 & 0.81\\
    \midrule
    \textbf{GPT-4o} & 0.68 & 0.57 & 0.49 & 0.32 & 0.23 & 0.21 & 0.11 & 0.37\\
    \textbf{GPT-4o-mini} & 0.63 & 0.42 & 0.34 & 0.17 & 0.09 & 0.10 & 0.01 & 0.25\\
    \textbf{NuminaMath-7B-CoT} & 0.28 & 0.13 & 0.12 & 0.05 & 0.01 & 0.00 & 0.00 & 0.08 \\
    \textbf{Deepseek-Math-7B} & 0.35 & 0.21 & 0.08 & 0.06 & 0.02 & 0.00 & 0.00 & 0.10\\
    \textbf{Qwen2.5-Base-7B} & 0.41 & 0.34 & 0.16 & 0.09 & 0.00 & 0.00 & 0.00 & 0.14 \\
    \midrule
    \textbf{Qwen2.5-7B-Instruct-1M} & 0.49 & 0.40 & 0.25 & 0.11 & 0.06 & 0.02 & 0.01 & 0.19 \\
    \rowcolor{bg!70}
    + \textbf{Logic-RL} & 0.99\bbonus{0.50} & 0.99\bbonus{0.59} & 0.94\bbonus{0.69} & 0.92\bbonus{0.81} & 0.91\bbonus{0.85} & 0.80\bbonus{0.78} & 0.67\bbonus{0.48} & 0.89\bbonus{0.70}\\
    \bottomrule
    \end{tabular}
    }
    \vspace{4mm}
    \caption{\centering Comparison of different models including reasoning models and general models on K\&K logic puzzle across various difficulty.}
    \label{tab:distill_vs_rl}
\end{table}

Compared to the initial average length of 500 tokens, after 1k steps of RL, the output has almost linearly and steadily increased to 2000 tokens, a significant increase of 4 times. As the response length increases, the model begins to exhibit more complex behaviors, such as reflection and exploration of alternative solutions. These behaviors emerge naturally, with no related data in our training set, and enhance the model's ability to handle more complex tasks. These phenomena align closely with the results of R1~\cite{deepseekai2025deepseekr1incentivizingreasoningcapability}. We will further discuss this in Section~\ref{rq3}.

\vspace{-2mm}

\section{Research Questions}

\subsection*{RQ 1: How Does GRPO Compare to Other RL Algorithms?}
\label{rl_algos}

\emph{Does GRPO\cite{grpo} outperform other reinforcement learning algorithms, such as REINFORCE++ and PPO, in terms of training stability, speed, and performance accuracy?}

% To ensure a fair comparison, we applied 3 algorithms solely to the advantage estimator while keeping all other components consistent. Additionally, we made minor modifications to the original implementations of PPO and REINFORCE++, following recommendations from Deepseek-Math~\cite{grpo}. These small adjustments were primarily motivated by considerations of efficiency and computational accuracy

\begin{figure}[H]
    \centering
    \includegraphics[width=1\linewidth]{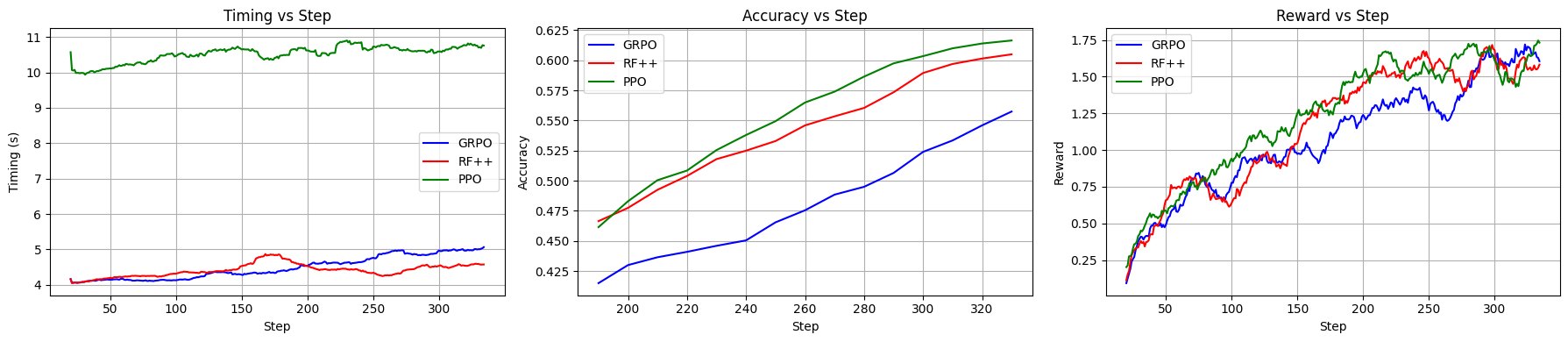}
    \caption{Comparison of GRPO (Blue), REINFORCE++ (Red), and PPO (Green) performance (averaged by sliding window = 50) in terms of training speed, accuracy, and reward gain.}
    \label{fig:rl}
\end{figure}

The results indicate that PPO~\cite{ppo} achieved significant advantages in both accuracy and reward. However, it was 138\% slower than REINFORCE++ in terms of training speed.
On the other hand, REINFORCE++~\cite{rpp} demonstrated superior stability, performance gains, and training efficiency compared to GRPO. Overall, REINFORCE++ outperformed GRPO across nearly all metrics, with \textbf{GRPO exhibiting the weakest performance among the three reinforcement learning algorithms evaluated in our experiments.}

\subsection*{RQ 2. Do certain thinking tokens and language-mixing phenemona improve reasoning?}
\emph{Does the inclusion of complex reasoning behaviours (such as exploration, verification, summarization, and backtracking) and language switching improve the model’s reasoning ability?}

\begin{figure}[H]
    \hspace{-0.5mm}\includegraphics[width=1\linewidth]{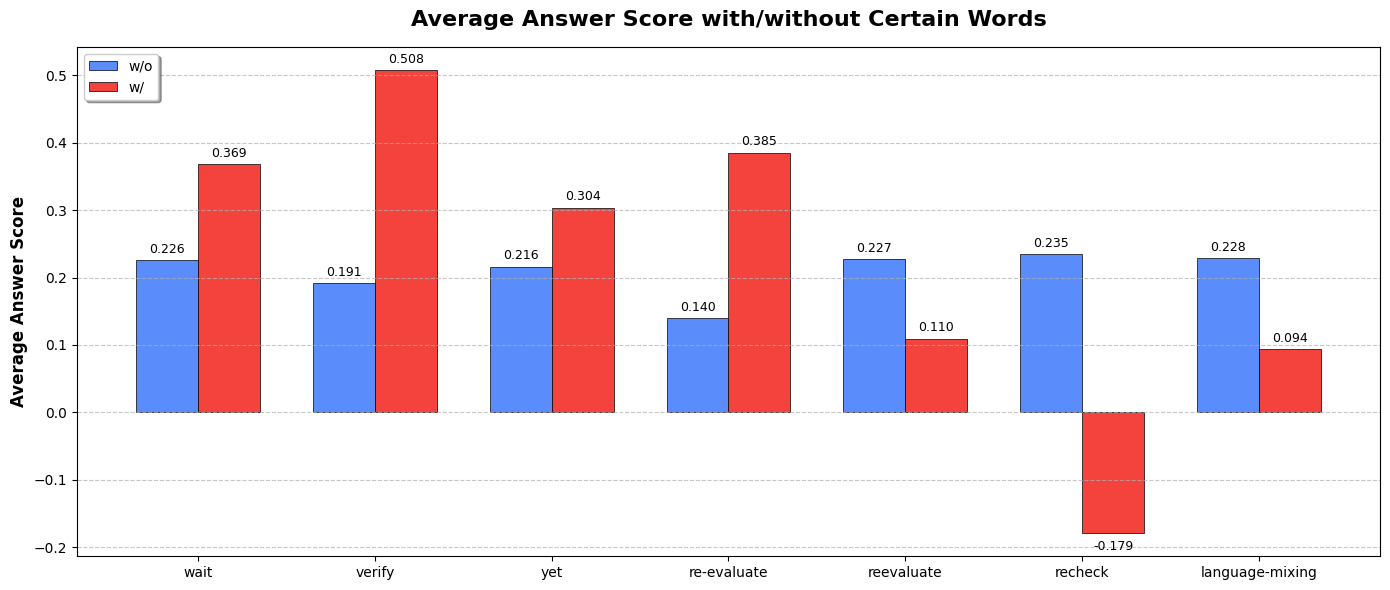}
    \caption{Impact of complex reasoning behaviours and language mixing on reasoning performance. We analyzed the model's answer rewards for responses containing the tokens shown in the figure. Responses with "verify" and "re-evaluate" scored significantly higher than those without these words. Conversely, responses containing certain tokens from other languages generally received lower scores.}
\end{figure}

We observe that in our experiments:
\begin{enumerate}[leftmargin=*]
    \item Language mixing significantly decreases reasoning ability. 
    % When Chinese or French words appear in responses to English queries, the model struggles to achieve a positive format score, let alone a positive answer reward.
    \item  While terms like "wait," "verify," "yet," and "re-evaluate" show significant improvement, not all complex thinking tokens enhance reasoning ability, as exemplified by "recheck."
    \item The complex reasoning behaviour “recheck” markedly diminishes reasoning ability, likely because its use signals the model’s uncertainty about its answer.
    \item There’s a clear difference between "re-evaluate" and "reevaluate": the former leads to much higher answer scores, while the latter lowers them. When we checked its origin responses, "reevaluate" almost never appeared, while "re-evaluate" showed up frequently. This may suggest the model is more comfortable with words it has seen more often in pretrain corpus.
\end{enumerate}

\subsection*{RQ 3: Does an 'Aha Moment' Emerge During Training?}
\emph{Is there an observable 'Aha moment' where the model exhibits a significant leap in reasoning capability, such as the emergence of multi-step verification or reflection during the RL process?}

The emergence of sophisticated behaviors becomes increasingly evident as performance grows. These behaviors include reflective actions—where the model revisits and reevaluates prior steps—and the spontaneous exploration of alternative problem-solving strategies. Such behaviors are not explicitly planted into training corpus but emerge organically through the model's interaction with the reinforcement learning environment, consistent with findings by Ye et al.~\cite{ye2024physicslanguagemodels21}.

\begin{figure}[H]
    \centering
    \captionsetup[subfigure]{labelformat=simple, labelsep=colon}
    
    % 第一行：三个子图（顶部对齐）
    \begin{subfigure}[t]{0.3\textwidth}
        \includegraphics[width=\textwidth, keepaspectratio]{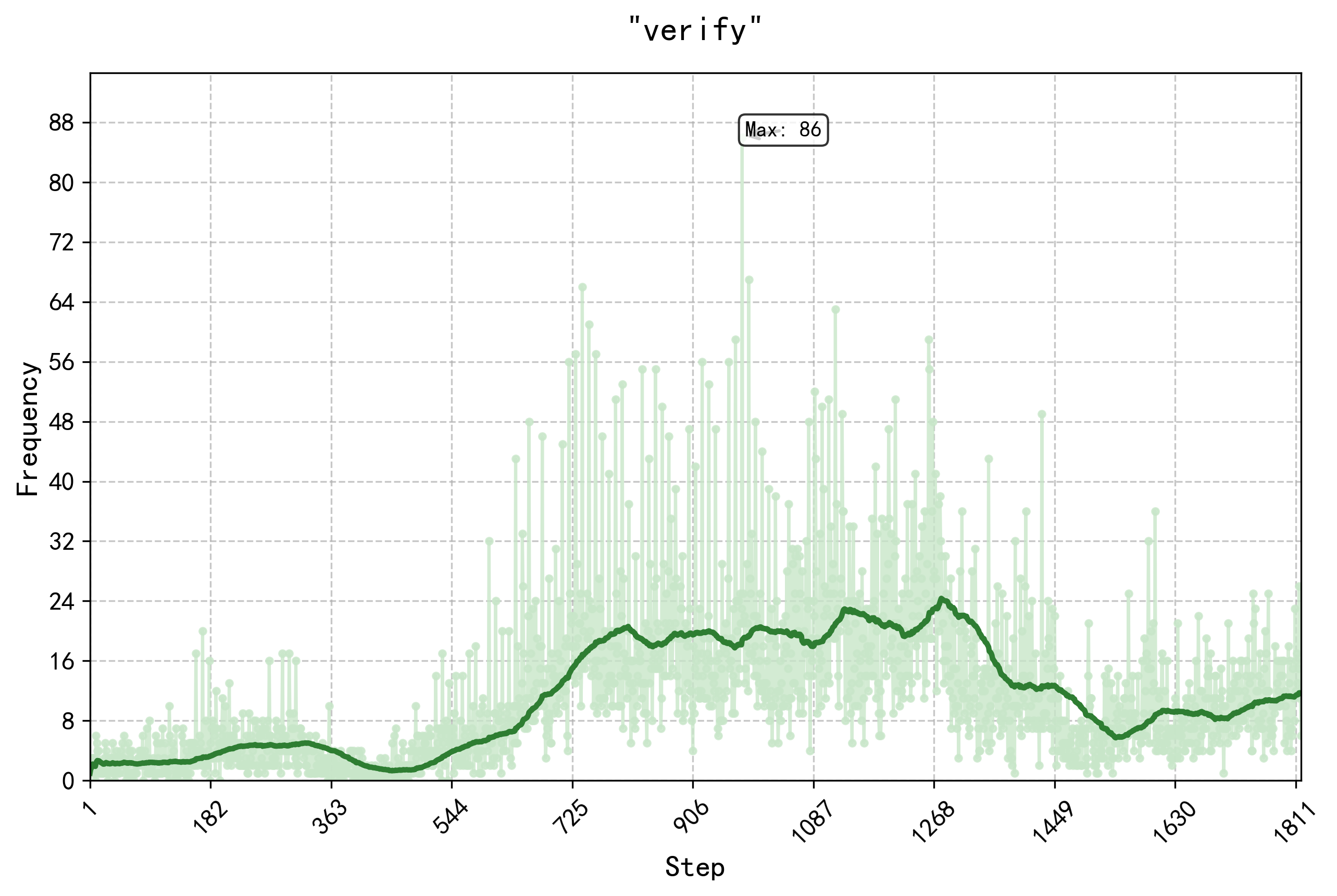}
        \subcaption{Verify}
        \label{fig:verify}
    \end{subfigure}
    \hspace{-0.01\textwidth} % 减少水平间距
    \begin{subfigure}[t]{0.3\textwidth}
        \includegraphics[width=\textwidth, keepaspectratio]{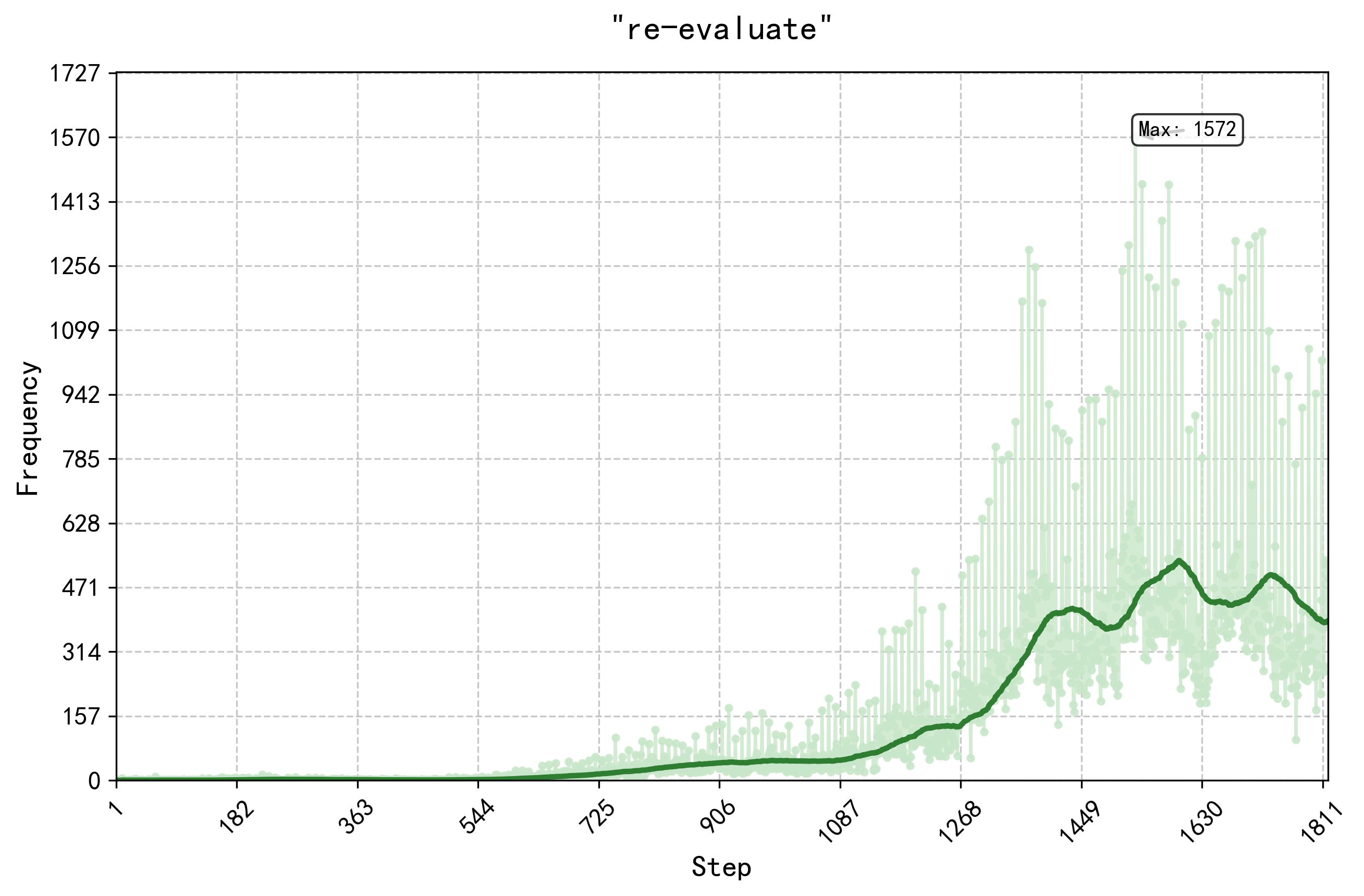}
        \subcaption{Re-evaluate}
        \label{fig:reevaluate}
    \end{subfigure}
    \hspace{-0.01\textwidth}
    \begin{subfigure}[t]{0.3\textwidth}
        \includegraphics[width=\textwidth, keepaspectratio]{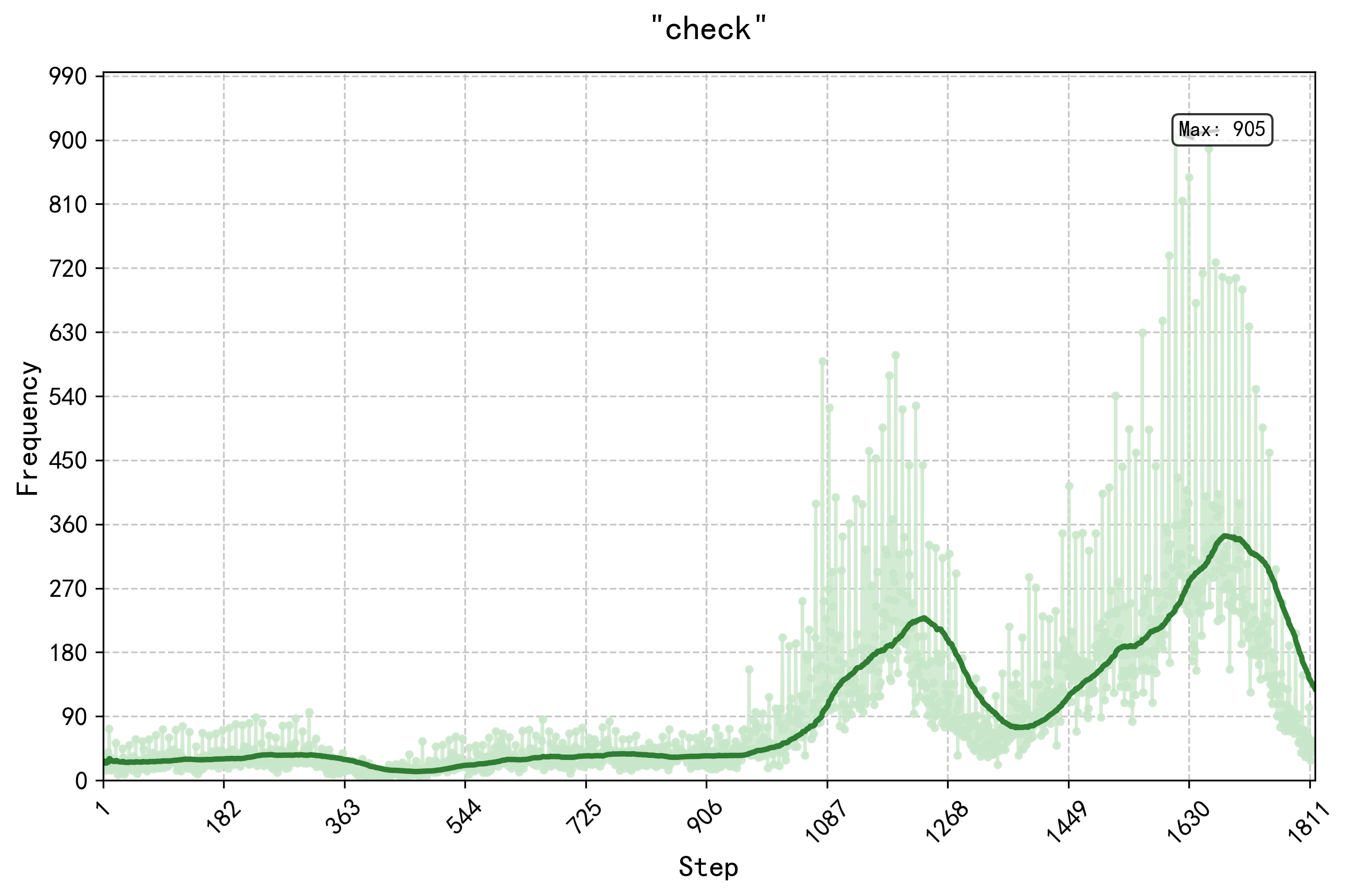}
        \subcaption{Check}
        \label{fig:check}
    \end{subfigure}
    
    \vspace{0.52em} % 减少垂直间距
    
    % 第二行：三个子图（顶部对齐）
    \begin{subfigure}[t]{0.3\textwidth}
        \includegraphics[width=\textwidth, keepaspectratio]{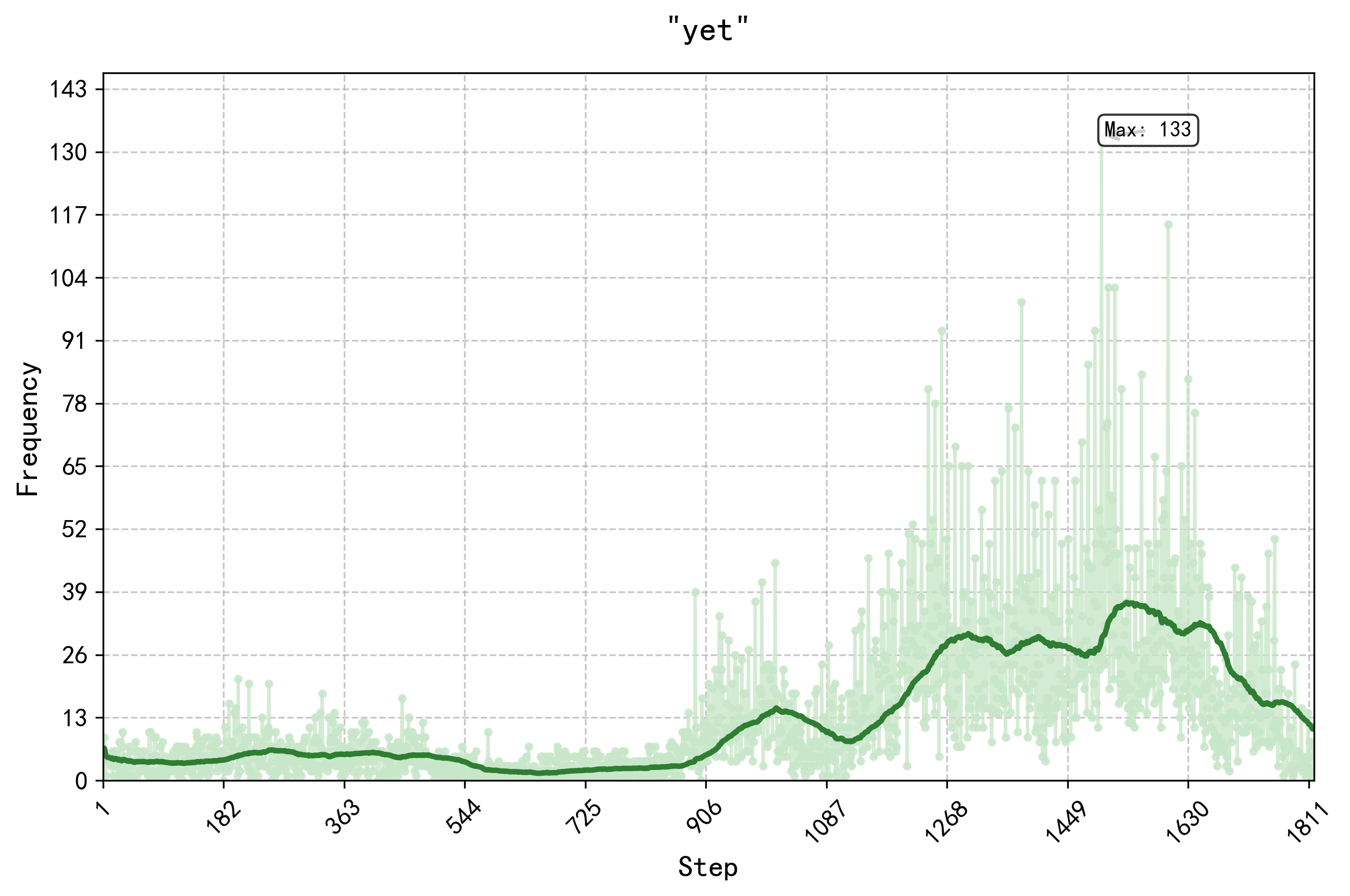}
        \subcaption{Yet}
        \label{fig:yet}
    \end{subfigure}
    \hspace{-0.01\textwidth}
    \begin{subfigure}[t]{0.3\textwidth}
        \includegraphics[width=\textwidth, keepaspectratio]{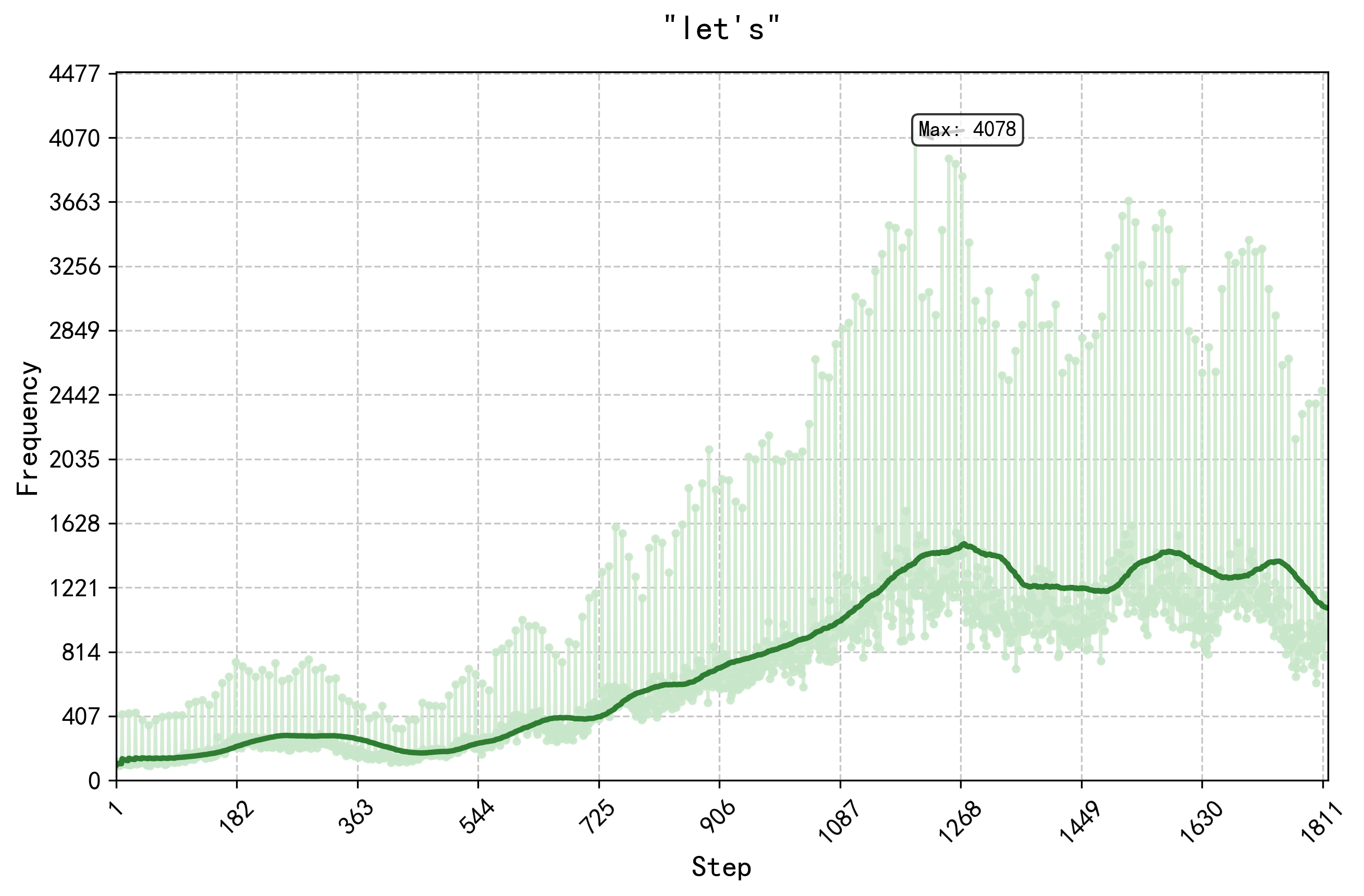}
        \subcaption{Let's}
        \label{fig:Let's}
    \end{subfigure}
    \hspace{-0.01\textwidth}
    \begin{subfigure}[t]{0.3\textwidth}
        \includegraphics[width=\textwidth, keepaspectratio]{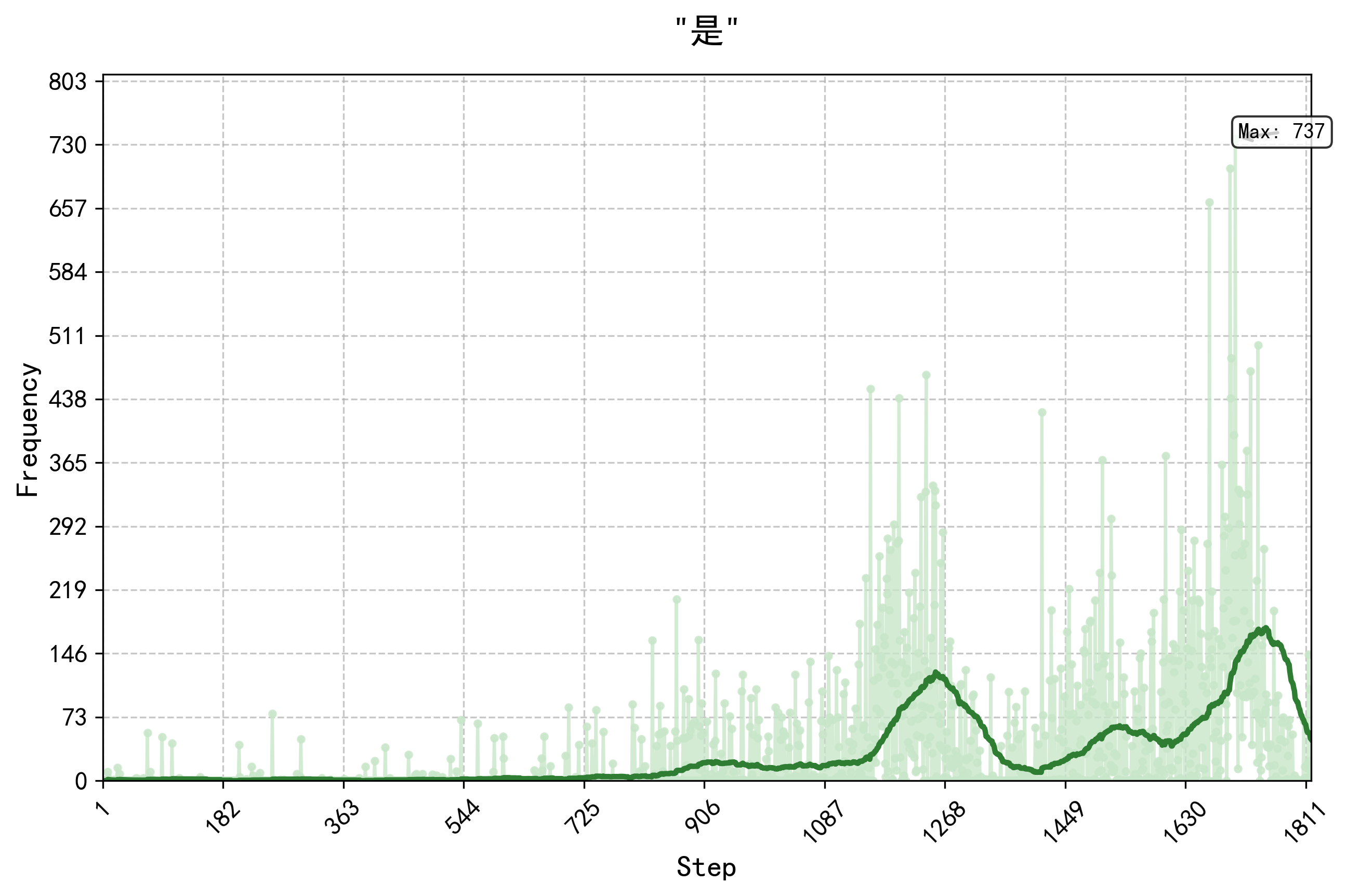}
        \subcaption{Chinese word}
        \label{fig:shi}
    \end{subfigure}
    
    \caption{Tracking the frequency of words in the first 1,800 training steps. \textbf{1.} Reflective words like "check" and "verify" slowly increased (a)-(c). \textbf{2.} Conversational phrases (e.g., "Let’s") and cautious terms (e.g., "yet") became more frequent  (d)-(e), \textbf{3.} Chinese words began appearing in English responses (f). The frequency of all these words developed steadily without sudden jumps,  suggesting that there may not be a distinct "aha moment."}
    \label{fig:keywords}
\end{figure}

The "aha moment" referenced in the R1 report~\cite{deepseekai2025deepseekr1incentivizingreasoningcapability} primarily refers to the model's sudden acquisition of complex reasoning behaviors. A secondary interpretation involves the model spontaneously verbalizing "aha moment," such as in phrases like "Wait, wait. Wait. That’s an aha moment I can flag here." While our model did not exhibit this specific verbalization, Figure~\ref{fig:keywords} shows that it displayed some complex reasoning behaviors (e.g., self-reflection, exploration, verification, summarization) even by step 10.

Thus, we conclude that \textbf{the RL process likely lacks a sudden "aha moment"}—that is, complex reasoning behaviors do not abruptly emerge at a specific training step, aligned with Liu et al.~\cite{liu2025oatzero}.

\subsection*{RQ 4: Can the Model Generalize to Out-of-Distribution (OOD) Tasks?}  
\emph{To what extent can the trained model handle tasks that differ from its training data, particularly those that are out-of-distribution?}

\begin{figure}[H] 
\centering
\begin{minipage}{0.48\linewidth}
    \centering
    \includegraphics[width=\linewidth]{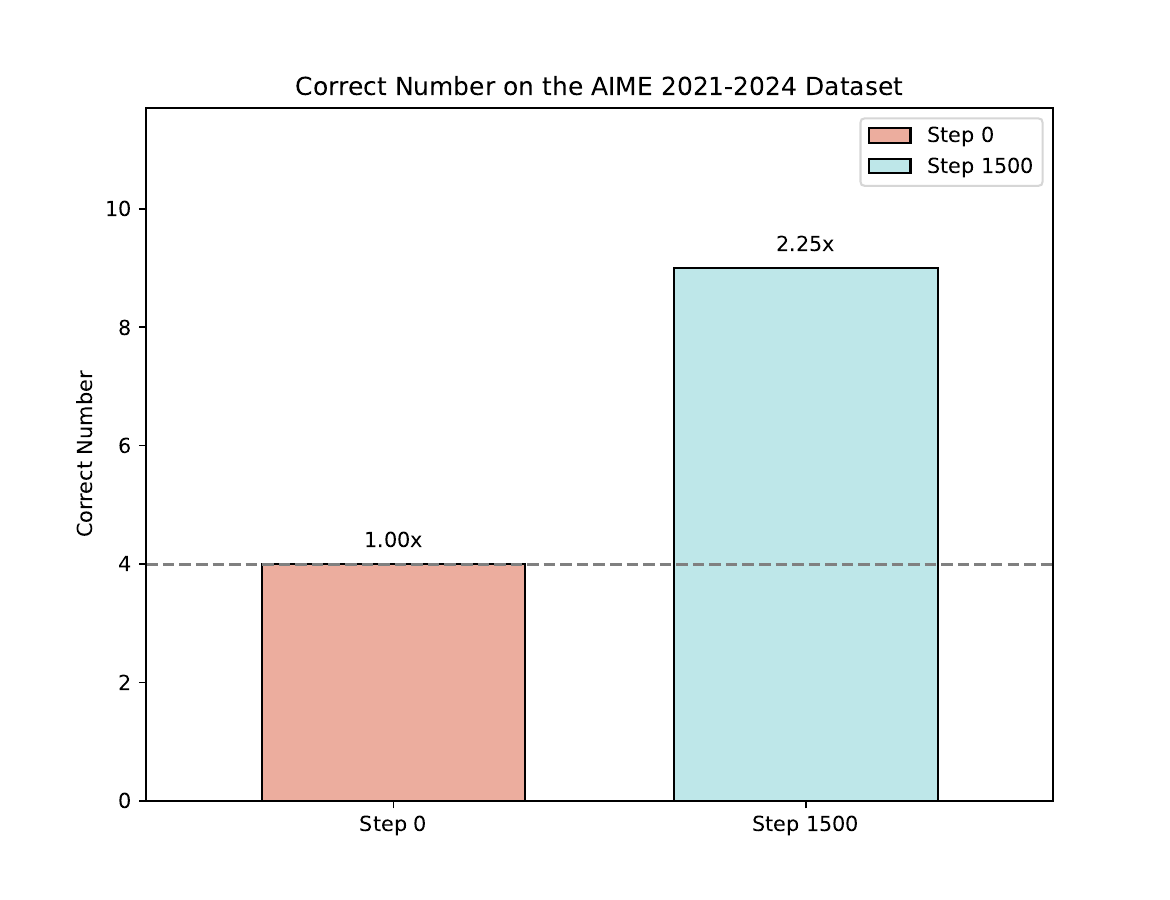} 
    \label{fig:aime}
\end{minipage}
\hspace{0.02\linewidth}
\begin{minipage}{0.48\linewidth}
    \centering
    \includegraphics[width=\linewidth]{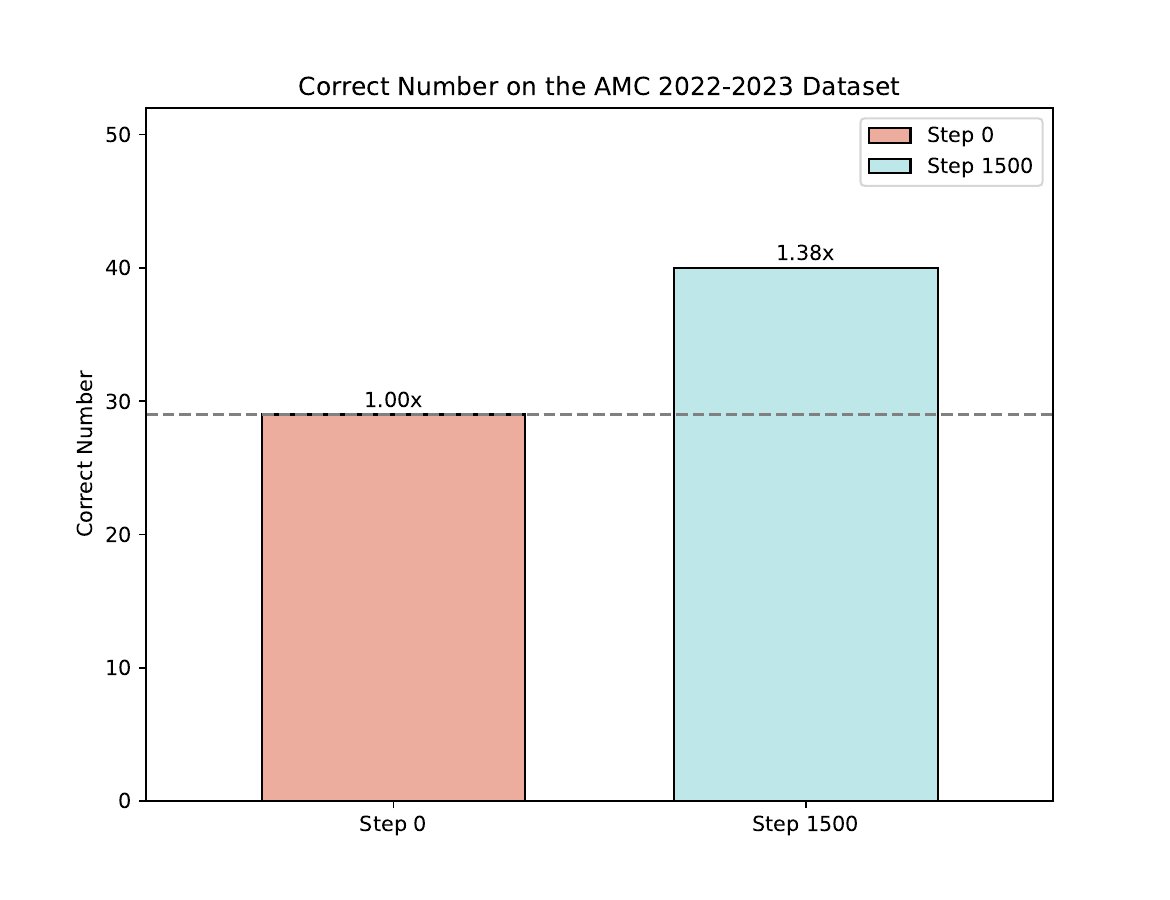} 
    \label{fig:amc}
\end{minipage}
\vspace{-4mm}
\caption{Training Step vs. Accuracy on AIME (2021-2024) and AMC (2022-2023) Datasets.}
\label{fig:ood}
\end{figure}
\vspace{-2mm}

The ability of a model to generalize beyond its training distribution is a cornerstone of AI research. We investigate whether the complex reasoning abilities—such as exploration, verification, summarization, and backtracking—that naturally emerged during RL process can transfer to a highly challenging mathematical reasoning scenario, which we term \emph{Super OOD} .

This evaluation leverages the widely-adopted AIME 2021–2024 (American Invitational Mathematics Examination) and AMC 2022–2023 (American Mathematics Competitions) benchmarks, both known for their rigorous and diverse problem sets.

From Figure~\ref{fig:ood}, we observe that the model's Super OOD generalization capability is exceptionally strong, achieving an overall improvement of 125\% on the AIME dataset and 38\% on the AMC dataset. This synchronous improvement indicates that the reinforcement learning process not only enhances the model's performance on in-distribution tasks but also facilitates the emergence of robust and transferable reasoning strategies.

These findings show that the reasoning skills learned during RL training go beyond the specific patterns of the K\&K dataset. This highlights RL's potential to generalize beyond their training environment.

\vspace{-3mm}

\subsection*{RQ 5: Which Generalizes Better, SFT or RL?}

\emph{Can post-training methods achieve more than just superficial alignment, which just learns  format patterns? Can SFT or RL actually learn to learn, effectively generalizing to other domains?}

We investigate whether models merely memorize training data or truly learn reasoning skills. Following the setup in \cite{memllm}, we test this by comparing performance on familiar problems versus slightly altered ones.Two signs of memorization:  High accuracy on seen problems, and low accuracy on slightly perturbed versions. So, we measure these using two metrics, denote model as $f$, dataset as $\mathcal{D}$.
\begin{enumerate}[leftmargin=*]
    \item \textbf{Accuracy on Observed Problems}: $\mathrm{Acc}(f; \mathcal{D})$, the model's accuracy on the training set ($\mathrm{Tr}$). 
    \item \textbf{Consistency Ratio}: $\mathrm{CR}(f; \mathcal{D})$, the ratio of correct solutions after small changes (perturbations) to those solved without changes. Perturbations preserve the puzzle's core principle and difficulty.
\end{enumerate}
The \textbf{Local Inconsistency-based Memorization Score} is defined as:
\[
\mathrm{LiMem}(f; \mathcal{D}) = \mathrm{Acc}(f; \mathcal{D}) \cdot (1 - \mathrm{CR}(f; \mathcal{D}))
\]
This score captures both memorization and sensitivity to changes. If a model's performance drops significantly when the problem format is altered, it likely hasn't learned the true reasoning skills required to solve similar but modified puzzles.We study two types of perturbations: (i) changing one person's statement to another bool logic expression, and (ii) reordering the statements between different people. Examples of perturbations are as follows:
 
\begin{tcolorbox}[
    colframe=purple!70!black, % 边框颜色：深紫色带一点黑色
    colback=purple!10!white, % 背景颜色：浅紫色
    coltitle=white, % 标题文字颜色：白色
    fonttitle=\bfseries, % 标题字体加粗
    title=Perturbation Examples\label{long_open_q}, % 标题内容
    % attach boxed title to top left={ % 将标题放置在左上角
    %     yshift=-2mm, % 垂直偏移
    %     xshift=2mm % 水平偏移
    % },
    % boxed title style={ % 标题框样式
    %     colframe=purple!70!black, % 标题框边框颜色
    %     colback=purple!70!black % 标题框背景颜色
    % },
    sharp corners, % 圆角设置为锐角
    boxrule=0.5mm, % 边框宽度
    % drop fuzzy shadow=gray!30 % 添加柔和阴影
]
\textbf{Original Problem}: \\
Zoey remarked, "Oliver is not a knight". Oliver stated, "Oliver is a knight if and only if Zoey is a knave". So who is a knight and who is a knave?
\\
\textbf{Statement Perturbation}: \\
Zoey remarked, "Oliver is a knight or Zoey is a knight". Oliver stated, "Oliver is a knight if and only if Zoey is a knave"

\textbf{Reorder Perturbation}: \\
Oliver stated, "Oliver is a knight if and only if Zoey is a knave". Zoey remarked, "Oliver is not a knight"

\end{tcolorbox}

% \begin{equation*}
% \mathcal{J}_{RFT}(\theta)= \mathbb{E}[q \sim P_{sft}(Q), o \sim \pi_{sft}(O|q)]\left( \frac{1}{|o|}\sum_{t=1}^{|o|} \mathbb{I}(o) \log \pi_\theta(o_{t} | q, o_{<t})\right).
% \end{equation*}

% \begin{figure}[H]
% \centering
% \includegraphics[width=1\linewidth]{figure/limem_rl.jpg}
% \caption{Accuracy and memorization scores of the model after reinforcement learning on datasets with 3, 5, and 7 people.}
% \label{fig:rq4_rl}
% \end{figure}

While it is challenging to collect a suitable SFT dataset, we employ reject sampling to gather ground truth data, referred to as the RFT method.

To explore which training paradigm offers better generalization performance, we apply two types of disturbances on training dataset, then compare RFT and RL with \textbf{Test acc - Mem Score curve}. For RFT settings, we use Reject Sampling on origin model, then use a rule-based Best-of-N method to collect the correct yet the shortest response for further fine-tune. As a result, we obtained the curve shown in Fig. 6, which illustrates the changes in unseen test accuracy and training dataset memorization over training process.

\begin{figure}[H]
\centering
\captionsetup[subfigure]{labelformat=simple, labelsep=colon}
% 左子图
\begin{subfigure}[t]{0.48\linewidth}
    \centering
    \includegraphics[width=\linewidth, keepaspectratio]{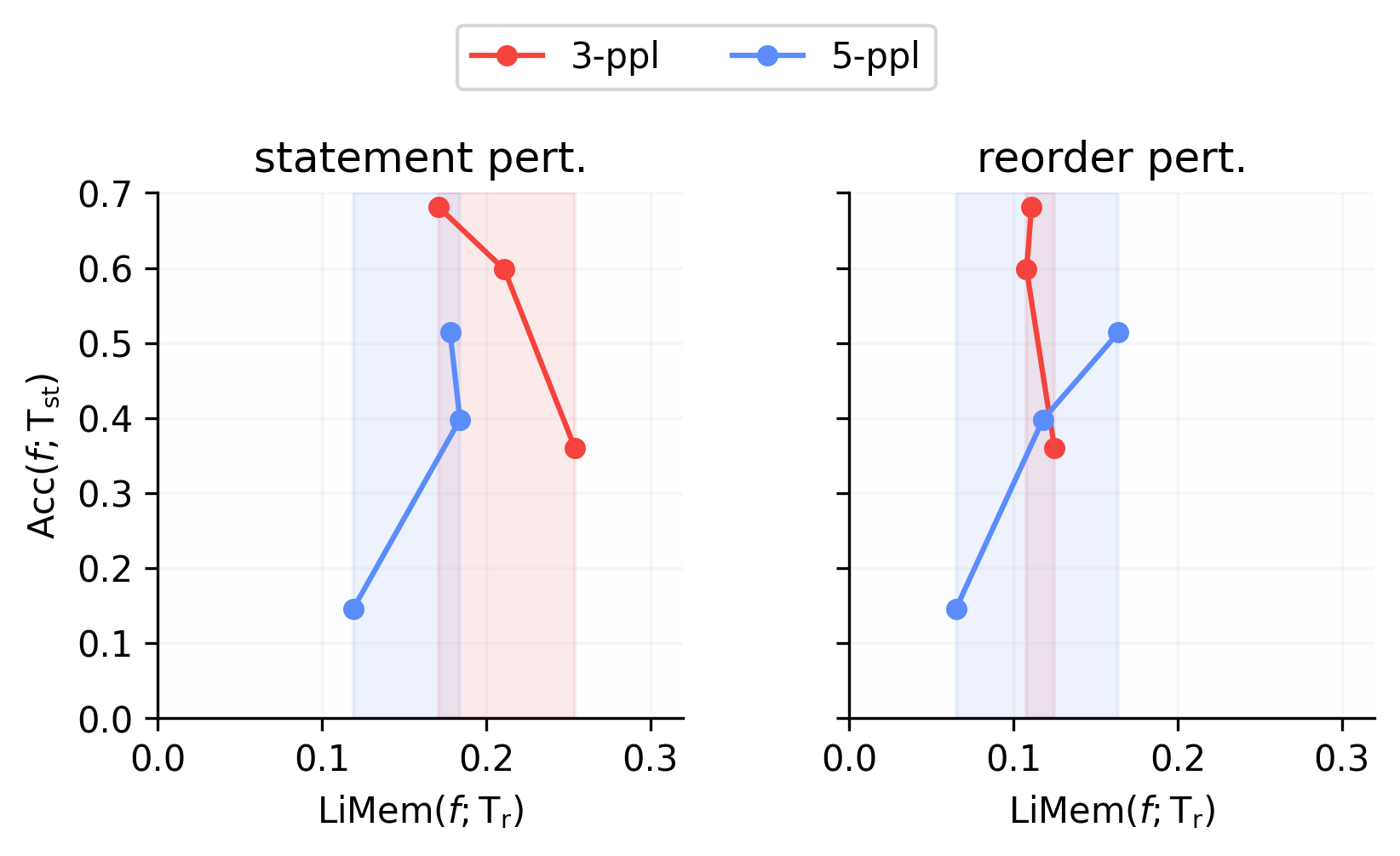} % 左子图文件名
    \subcaption{RL}
    \label{fig:rq4_rl}
\end{subfigure}
\hfill
% 右子图
\begin{subfigure}[t]{0.48\linewidth}
    \centering
    \includegraphics[width=\linewidth, keepaspectratio]{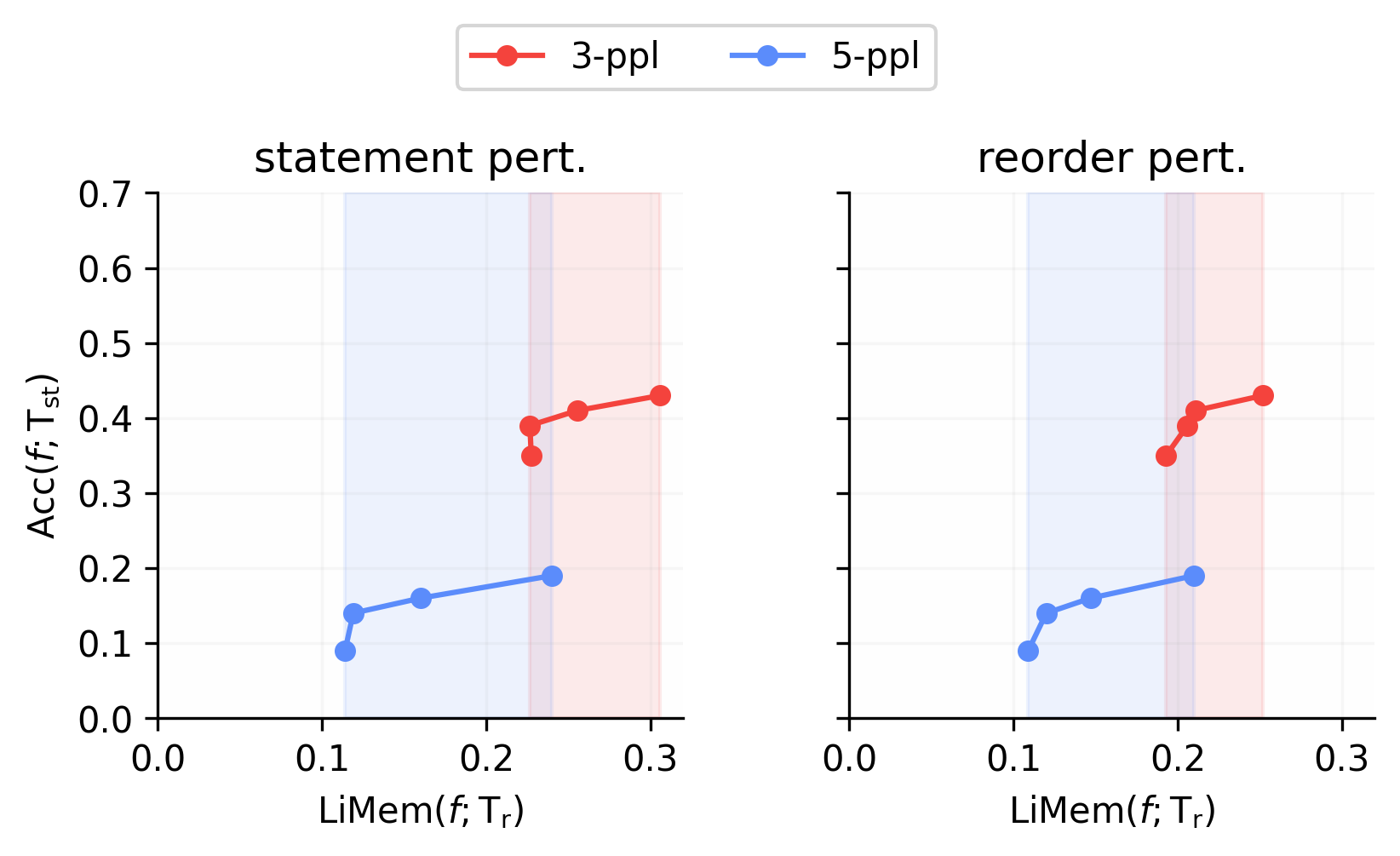} % 右子图文件名
    \subcaption{RFT}
    \label{fig:rq4_rft}
\end{subfigure}
% 主标题
\caption{\textbf{RFT memorizes while RL generalizes.} \textbf{RFT} (Reject sampling Fine-Tuning) slightly improves test accuracy at the expense of rapidly increasing $LiMem(f;Tr)$, indicating it mainly learns superficial answer format than geniue reasoning. In contrast, \textbf{RL}  achieves higher test accuracy with minimal or even negative increase in $LiMem(f;Tr)$. Within the same $LiMem$ interval, RL outperform RFT in test acc greatly, suggesting better generalization ability. }
\label{fig:rq4_rl}
\end{figure}

A higher memorization score indicates that more questions, which were originally correct, have turned incorrect due to the disturbances caused by the training, reflecting a greater degree of memorization. SFT tends to superficial alignment, often becoming overly reliant on the original data's expression format. On the other hand, RL encourages the model to explore independently, fostering generalization abilities that stem from enhanced reasoning capabilities, consistent with findings in \cite{chu2025sftmemorizesrlgeneralizes}.

\subsection*{RQ 6: Is Curriculum Learning Still Necessary in RL?}
\emph{Does curriculum learning still matter in the RL paradigm? Specifically, is the sequential order of data important, or is it merely the curation ratio that matters?}
\begin{figure}[H]
\centering
\includegraphics[width=1\linewidth]{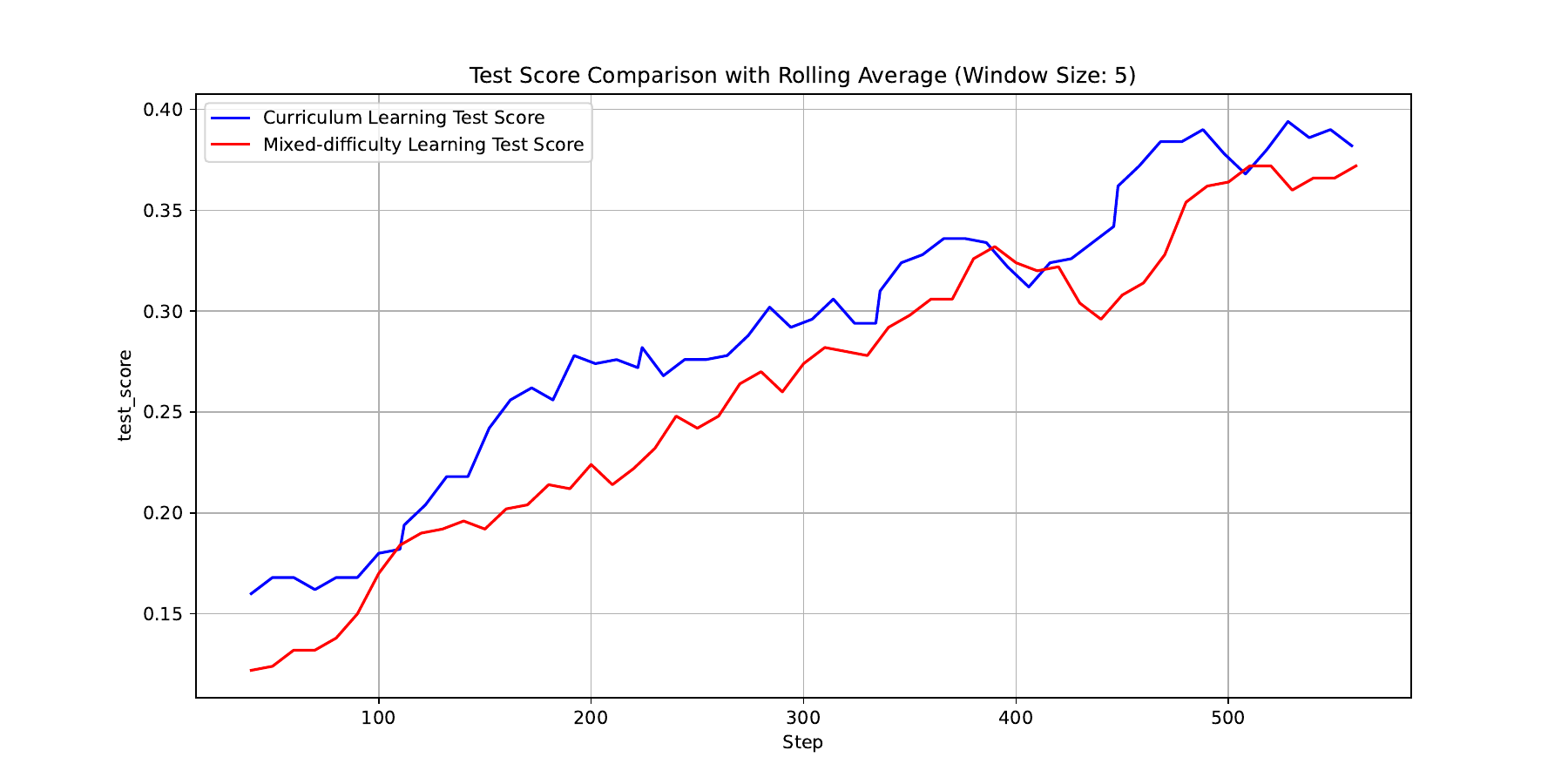}
\caption{Comparison of test scores for curriculum learning and mixed-difficulty training. The plot uses a rolling average (window size = 5)}
\label{fig:rq5_curriculum}
\end{figure}
% To evaluate the necessity of curriculum learning in our framework, we compare its effectiveness against a mixed-difficulty training approach. This comparison aims to verify whether the theoretically appealing concept of curriculum learning—gradually exposing models to increasingly complex tasks—provides tangible benefits over directly exposing them to a diverse range of difficulty levels. In our implementation, the curriculum learning strategy sequentially trains the model on datasets ranging from 3-7 people scenarios for one epoch each, with each training set representing an ascending level of difficulty. In contrast, the mixed-difficulty approach trains the model simultaneously on all difficulty levels (3–7 people) within a single epoch. All other hyperparameters are kept identical between the two training regimes.

To evaluate the necessity of curriculum learning, we compare its effectiveness to a mixed-difficulty approach. In curriculum learning, the model is trained sequentially on progressively more difficult datasets (3-7 people scenarios) for one epoch each. In contrast, the mixed-difficulty approach trains the model on all difficulty levels simultaneously within a single epoch. All other hyperparameters are kept constant between the two methods.

We analyze the test score trajectories using a rolling average (window size = 5) to mitigate stochastic fluctuations and highlight underlying trends. The results in Figure \ref{fig:rq5_curriculum} indicate that curriculum learning yields slightly higher test scores during intermediate training phases. However, this advantage diminishes in practical significance, as the performance difference during early training stages remains statistically negligible, suggesting a limited impact on initial convergence. While curriculum learning may offer a marginal theoretical benefit in terms of sample efficiency, its practical necessity is not conclusively supported, given the minimal real-world performance difference and the added complexity of staged training.

\subsection*{RQ 7: Does Longer Response Length Guarantee Better Reasoning?}
\label{rq3}

\emph{Does an increase in response length during training directly improve a model's reasoning performance, or are these trends merely correlated?}

\begin{figure}[H]
\centering
\includegraphics[width=1\linewidth]{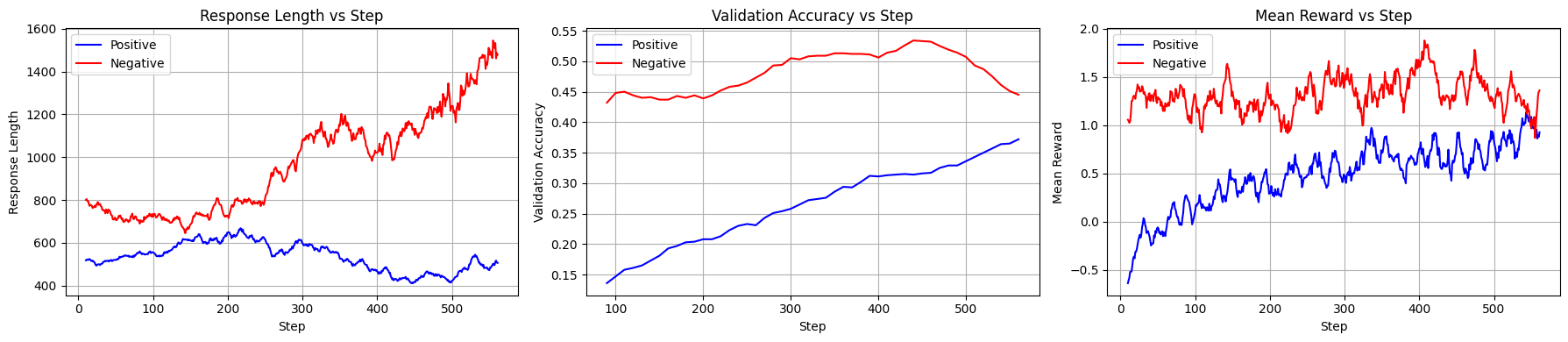}
\caption{Comparison of response length, validation accuracy, and mean reward across training steps for positive and negative example models.}
\label{fig:rq3}
\end{figure}

This experiment investigates whether an increase in response length during training causally enhances reasoning performance. We compared two models trained using the same algorithm and base model but with different hyperparameters and dataset difficulties.

\begin{itemize}[leftmargin=*]
\item \textbf{Positive Example Model (Blue):} Despite a slight decrease in response length over time, this model demonstrated significant improvements in both validation accuracy and reward, indicating stronger reasoning and generalization abilities.

\item \textbf{Negative Example Model (Red):} This model consistently increased its response length but showed no improvement in validation accuracy or reward. This suggests that increasing response length alone does not necessarily enhance reasoning capabilities.
\end{itemize}

Figure \ref{fig:rq3} illustrates these findings: the positive model's reward and accuracy improved while its response length decreased, whereas the negative model's length increased without any corresponding performance gains. These divergent trends suggest that changes in response length are likely a byproduct of training dynamics rather than a causal driver of reasoning improvements.

The observed increase in response length is likely a side effect of reinforcement learning (RL) dynamics. While some studies report a natural tendency for output length to grow as models generate more complex responses, this growth should be viewed as a correlate rather than a direct cause of improved reasoning. Importantly, there is no statistically significant evidence that the magnitude of length increase reliably predicts proportional gains in reasoning performance.

In conclusion, \textbf{longer responses do not always guarantee better reasoning.} Enhanced reasoning capabilities may naturally lead to more detailed explanations, but artificially extending response length does not necessarily result in proportional performance improvements.

\section{Discussion and Future Work}
While our study demonstrates the potential of Logic-RL in developing complex reasoning skills, it is important to note that our findings are based on a small-scale logic dataset. The generalizability of our results to large-scale real-world mathematical or coding scenarios remains to be explored. Future work should focus on extending our approach to more diverse and complex datasets to further validate its effectiveness and robustness. Our work will remain an \textbf{open research project} to benefit the community.

\paragraph{Chain-of-Thought Long to Short Methods.}

Despite the fact that our initial prompt is concise and straightforward, the length of the responses can expand by up to four times after reinforcement learning training. In order to enhance token efficiency and accommodate a long-context friendly training paradigm, we find it particularly valuable to explore methods that transform long responses into shorter, more digestible formats. This investigation aims at improving overall efficiency and effectiveness in handling lengthy outputs, thereby optimizing the training process for better scalability and performance.

\paragraph{Stablize RL Training.}

We have found it beneficial in some cases to eliminate KL constraints, especially when starting from a strong foundation model. Additionally, introducing a higher temperature at the beginning of training appears to provide the model with a more diverse starting point. We plan to further investigate how the SFT stage impacts the effectiveness and efficiency of RL training.

\paragraph{Mixed-Language Reasoning.} 

A curious phenomenon is the model’s frequent use of Chinese tokens in the \texttt{<think>} section, despite training data being fully in English. One hypothesis is that certain tokens in the Chinese vocabulary vector might produce hidden states that are ``favorable'' under our RL scheme. Investigating whether code-switching or even \emph{random token switching} could systematically aid internal reasoning is an exciting avenue.

\paragraph{Relaxing the Formatting Constraints.}

Although \texttt{<think>} \ldots \texttt{</think>} effectively organizes the chain of thought, it remains an open question whether an entirely unconstrained or latent approach might yield better results. The model might eventually ``invent'' its own internal representation for reasoning if given the right incentives.

\bibliography{references}{}

\begin{thebibliography}{10}

\bibitem{chu2025sftmemorizesrlgeneralizes}
Tianzhe Chu, Yuexiang Zhai, Jihan Yang, Shengbang Tong, Saining Xie, Dale Schuurmans, Quoc~V. Le, Sergey Levine, and Yi~Ma.
\newblock Sft memorizes, rl generalizes: A comparative study of foundation model post-training, 2025.

\bibitem{cobbe2021trainingverifierssolvemath}
Karl Cobbe, Vineet Kosaraju, Mohammad Bavarian, Mark Chen, Heewoo Jun, Lukasz Kaiser, Matthias Plappert, Jerry Tworek, Jacob Hilton, Reiichiro Nakano, Christopher Hesse, and John Schulman.
\newblock Training verifiers to solve math word problems, 2021.

\bibitem{deepseekai2025deepseekr1incentivizingreasoningcapability}
DeepSeek-AI, Daya Guo, Dejian Yang, Haowei Zhang, Junxiao Song, Ruoyu Zhang, Runxin Xu, Qihao Zhu, Shirong Ma, Peiyi Wang, Xiao Bi, Xiaokang Zhang, Xingkai Yu, Yu~Wu, Z.~F. Wu, Zhibin Gou, Zhihong Shao, Zhuoshu Li, Ziyi Gao, Aixin Liu, Bing Xue, Bingxuan Wang, Bochao Wu, Bei Feng, Chengda Lu, Chenggang Zhao, Chengqi Deng, Chenyu Zhang, Chong Ruan, Damai Dai, Deli Chen, Dongjie Ji, Erhang Li, Fangyun Lin, Fucong Dai, Fuli Luo, Guangbo Hao, Guanting Chen, Guowei Li, H.~Zhang, Han Bao, Hanwei Xu, Haocheng Wang, Honghui Ding, Huajian Xin, Huazuo Gao, Hui Qu, Hui Li, Jianzhong Guo, Jiashi Li, Jiawei Wang, Jingchang Chen, Jingyang Yuan, Junjie Qiu, Junlong Li, J.~L. Cai, Jiaqi Ni, Jian Liang, Jin Chen, Kai Dong, Kai Hu, Kaige Gao, Kang Guan, Kexin Huang, Kuai Yu, Lean Wang, Lecong Zhang, Liang Zhao, Litong Wang, Liyue Zhang, Lei Xu, Leyi Xia, Mingchuan Zhang, Minghua Zhang, Minghui Tang, Meng Li, Miaojun Wang, Mingming Li, Ning Tian, Panpan Huang, Peng Zhang, Qiancheng Wang, Qinyu Chen, Qiushi Du, Ruiqi Ge, Ruisong
  Zhang, Ruizhe Pan, Runji Wang, R.~J. Chen, R.~L. Jin, Ruyi Chen, Shanghao Lu, Shangyan Zhou, Shanhuang Chen, Shengfeng Ye, Shiyu Wang, Shuiping Yu, Shunfeng Zhou, Shuting Pan, S.~S. Li, Shuang Zhou, Shaoqing Wu, Shengfeng Ye, Tao Yun, Tian Pei, Tianyu Sun, T.~Wang, Wangding Zeng, Wanjia Zhao, Wen Liu, Wenfeng Liang, Wenjun Gao, Wenqin Yu, Wentao Zhang, W.~L. Xiao, Wei An, Xiaodong Liu, Xiaohan Wang, Xiaokang Chen, Xiaotao Nie, Xin Cheng, Xin Liu, Xin Xie, Xingchao Liu, Xinyu Yang, Xinyuan Li, Xuecheng Su, Xuheng Lin, X.~Q. Li, Xiangyue Jin, Xiaojin Shen, Xiaosha Chen, Xiaowen Sun, Xiaoxiang Wang, Xinnan Song, Xinyi Zhou, Xianzu Wang, Xinxia Shan, Y.~K. Li, Y.~Q. Wang, Y.~X. Wei, Yang Zhang, Yanhong Xu, Yao Li, Yao Zhao, Yaofeng Sun, Yaohui Wang, Yi~Yu, Yichao Zhang, Yifan Shi, Yiliang Xiong, Ying He, Yishi Piao, Yisong Wang, Yixuan Tan, Yiyang Ma, Yiyuan Liu, Yongqiang Guo, Yuan Ou, Yuduan Wang, Yue Gong, Yuheng Zou, Yujia He, Yunfan Xiong, Yuxiang Luo, Yuxiang You, Yuxuan Liu, Yuyang Zhou, Y.~X. Zhu,
  Yanhong Xu, Yanping Huang, Yaohui Li, Yi~Zheng, Yuchen Zhu, Yunxian Ma, Ying Tang, Yukun Zha, Yuting Yan, Z.~Z. Ren, Zehui Ren, Zhangli Sha, Zhe Fu, Zhean Xu, Zhenda Xie, Zhengyan Zhang, Zhewen Hao, Zhicheng Ma, Zhigang Yan, Zhiyu Wu, Zihui Gu, Zijia Zhu, Zijun Liu, Zilin Li, Ziwei Xie, Ziyang Song, Zizheng Pan, Zhen Huang, Zhipeng Xu, Zhongyu Zhang, and Zhen Zhang.
\newblock Deepseek-r1: Incentivizing reasoning capability in llms via reinforcement learning, 2025.

\bibitem{feng2024alphazeroliketreesearchguidelarge}
Xidong Feng, Ziyu Wan, Muning Wen, Stephen~Marcus McAleer, Ying Wen, Weinan Zhang, and Jun Wang.
\newblock Alphazero-like tree-search can guide large language model decoding and training, 2024.

\bibitem{gao2024omnimathuniversalolympiadlevel}
Bofei Gao, Feifan Song, Zhe Yang, Zefan Cai, Yibo Miao, Qingxiu Dong, Lei Li, Chenghao Ma, Liang Chen, Runxin Xu, Zhengyang Tang, Benyou Wang, Daoguang Zan, Shanghaoran Quan, Ge~Zhang, Lei Sha, Yichang Zhang, Xuancheng Ren, Tianyu Liu, and Baobao Chang.
\newblock Omni-math: A universal olympiad level mathematic benchmark for large language models, 2024.

\bibitem{sc-mcts}
Zitian Gao, Boye Niu, Xuzheng He, Haotian Xu, Hongzhang Liu, Aiwei Liu, Xuming Hu, and Lijie Wen.
\newblock Interpretable contrastive monte carlo tree search reasoning, 2024.

\bibitem{rpp}
Jian Hu.
\newblock Reinforce++: A simple and efficient approach for aligning large language models, 2025.

\bibitem{lightman2023letsverifystepstep}
Hunter Lightman, Vineet Kosaraju, Yura Burda, Harri Edwards, Bowen Baker, Teddy Lee, Jan Leike, John Schulman, Ilya Sutskever, and Karl Cobbe.
\newblock Let's verify step by step, 2023.

\bibitem{liu2025oatzero}
Zichen Liu, Changyu Chen, Wenjun Li, Tianyu Pang, Chao Du, and Min Lin.
\newblock There may not be aha moment in r1-zero-like training — a pilot study.
\newblock \url{https://oatllm.notion.site/oat-zero}, 2025.
\newblock Notion Blog.

\bibitem{o1}
OpenAI.
\newblock Introducing openai o1.
\newblock \url{https://openai.com/o1/}, 2024.
\newblock Accessed: 2024-10-02.

\bibitem{dpo}
Rafael Rafailov, Archit Sharma, Eric Mitchell, Christopher~D Manning, Stefano Ermon, and Chelsea Finn.
\newblock Direct preference optimization: Your language model is secretly a reward model.
\newblock In A.~Oh, T.~Naumann, A.~Globerson, K.~Saenko, M.~Hardt, and S.~Levine, editors, {\em Advances in Neural Information Processing Systems}, volume~36, pages 53728--53741. Curran Associates, Inc., 2023.

\bibitem{ppo}
John Schulman, Filip Wolski, Prafulla Dhariwal, Alec Radford, and Oleg Klimov.
\newblock Proximal policy optimization algorithms, 2017.

\bibitem{grpo}
Zhihong Shao, Peiyi Wang, Qihao Zhu, Runxin Xu, Junxiao Song, Xiao Bi, Haowei Zhang, Mingchuan Zhang, Y.~K. Li, Y.~Wu, and Daya Guo.
\newblock Deepseekmath: Pushing the limits of mathematical reasoning in open language models, 2024.

\bibitem{alphazero}
David Silver, Thomas Hubert, Julian Schrittwieser, Ioannis Antonoglou, Matthew Lai, Arthur Guez, Marc Lanctot, Laurent Sifre, Dharshan Kumaran, Thore Graepel, Timothy Lillicrap, Karen Simonyan, and Demis Hassabis.
\newblock Mastering chess and shogi by self-play with a general reinforcement learning algorithm, 2017.

\bibitem{kimiteam2025kimik15scalingreinforcement}
Kimi Team, Angang Du, Bofei Gao, Bowei Xing, Changjiu Jiang, Cheng Chen, Cheng Li, Chenjun Xiao, Chenzhuang Du, Chonghua Liao, Chuning Tang, Congcong Wang, Dehao Zhang, Enming Yuan, Enzhe Lu, Fengxiang Tang, Flood Sung, Guangda Wei, Guokun Lai, Haiqing Guo, Han Zhu, Hao Ding, Hao Hu, Hao Yang, Hao Zhang, Haotian Yao, Haotian Zhao, Haoyu Lu, Haoze Li, Haozhen Yu, Hongcheng Gao, Huabin Zheng, Huan Yuan, Jia Chen, Jianhang Guo, Jianlin Su, Jianzhou Wang, Jie Zhao, Jin Zhang, Jingyuan Liu, Junjie Yan, Junyan Wu, Lidong Shi, Ling Ye, Longhui Yu, Mengnan Dong, Neo Zhang, Ningchen Ma, Qiwei Pan, Qucheng Gong, Shaowei Liu, Shengling Ma, Shupeng Wei, Sihan Cao, Siying Huang, Tao Jiang, Weihao Gao, Weimin Xiong, Weiran He, Weixiao Huang, Wenhao Wu, Wenyang He, Xianghui Wei, Xianqing Jia, Xingzhe Wu, Xinran Xu, Xinxing Zu, Xinyu Zhou, Xuehai Pan, Y.~Charles, Yang Li, Yangyang Hu, Yangyang Liu, Yanru Chen, Yejie Wang, Yibo Liu, Yidao Qin, Yifeng Liu, Ying Yang, Yiping Bao, Yulun Du, Yuxin Wu, Yuzhi Wang, Zaida Zhou,
  Zhaoji Wang, Zhaowei Li, Zhen Zhu, Zheng Zhang, Zhexu Wang, Zhilin Yang, Zhiqi Huang, Zihao Huang, Ziyao Xu, and Zonghan Yang.
\newblock Kimi k1.5: Scaling reinforcement learning with llms, 2025.

\bibitem{cot}
Jason Wei, Xuezhi Wang, Dale Schuurmans, Maarten Bosma, brian ichter, Fei Xia, Ed~Chi, Quoc~V Le, and Denny Zhou.
\newblock Chain-of-thought prompting elicits reasoning in large language models.
\newblock In S.~Koyejo, S.~Mohamed, A.~Agarwal, D.~Belgrave, K.~Cho, and A.~Oh, editors, {\em Advances in Neural Information Processing Systems}, volume~35, pages 24824--24837. Curran Associates, Inc., 2022.

\bibitem{memllm}
Chulin Xie, Yangsibo Huang, Chiyuan Zhang, Da~Yu, Xinyun Chen, Bill~Yuchen Lin, Bo~Li, Badih Ghazi, and Ravi Kumar.
\newblock On memorization of large language models in logical reasoning, 2024.

\bibitem{deepprover1.5}
Huajian Xin, Z.~Z. Ren, Junxiao Song, Zhihong Shao, Wanjia Zhao, Haocheng Wang, Bo~Liu, Liyue Zhang, Xuan Lu, Qiushi Du, Wenjun Gao, Qihao Zhu, Dejian Yang, Zhibin Gou, Z.~F. Wu, Fuli Luo, and Chong Ruan.
\newblock Deepseek-prover-v1.5: Harnessing proof assistant feedback for reinforcement learning and monte-carlo tree search, 2024.

\bibitem{xu2025largereasoningmodelssurvey}
Fengli Xu, Qianyue Hao, Zefang Zong, Jingwei Wang, Yunke Zhang, Jingyi Wang, Xiaochong Lan, Jiahui Gong, Tianjian Ouyang, Fanjin Meng, Chenyang Shao, Yuwei Yan, Qinglong Yang, Yiwen Song, Sijian Ren, Xinyuan Hu, Yu~Li, Jie Feng, Chen Gao, and Yong Li.
\newblock Towards large reasoning models: A survey of reinforced reasoning with large language models, 2025.

\bibitem{xu2023traingainunleashmathematical}
Haotian Xu.
\newblock No train still gain. unleash mathematical reasoning of large language models with monte carlo tree search guided by energy function, 2023.

\bibitem{xu2025redstardoesscalinglongcot}
Haotian Xu, Xing Wu, Weinong Wang, Zhongzhi Li, Da~Zheng, Boyuan Chen, Yi~Hu, Shijia Kang, Jiaming Ji, Yingying Zhang, Zhijiang Guo, Yaodong Yang, Muhan Zhang, and Debing Zhang.
\newblock Redstar: Does scaling long-cot data unlock better slow-reasoning systems?, 2025.

\bibitem{qwen251m}
An~Yang, Bowen Yu, Chengyuan Li, Dayiheng Liu, Fei Huang, Haoyan Huang, Jiandong Jiang, Jianhong Tu, Jianwei Zhang, Jingren Zhou, Junyang Lin, Kai Dang, Kexin Yang, Le~Yu, Mei Li, Minmin Sun, Qin Zhu, Rui Men, Tao He, Weijia Xu, Wenbiao Yin, Wenyuan Yu, Xiafei Qiu, Xingzhang Ren, Xinlong Yang, Yong Li, Zhiying Xu, and Zipeng Zhang.
\newblock Qwen2.5-1m technical report, 2025.

\bibitem{ye2024physicslanguagemodels21}
Tian Ye, Zicheng Xu, Yuanzhi Li, and Zeyuan Allen-Zhu.
\newblock Physics of language models: Part 2.1, grade-school math and the hidden reasoning process, 2024.

\bibitem{workingmemory}
Chunhui Zhang, Yiren Jian, Zhongyu Ouyang, and Soroush Vosoughi.
\newblock Working memory identifies reasoning limits in language models.
\newblock In Yaser Al-Onaizan, Mohit Bansal, and Yun-Nung Chen, editors, {\em Proceedings of the 2024 Conference on Empirical Methods in Natural Language Processing}, pages 16896--16922, Miami, Florida, USA, November 2024. Association for Computational Linguistics.

\end{thebibliography}
\bibliographystyle{plain}

\renewcommand{\thesubsection}{\Alph{subsection}}
\section*{Appendix}
\setcounter{subsection}{0}
\section{Related Work}
\paragraph{Large Language Model Reasoning}
A key focus for LLMs is improving their reasoning abilities, particularly for complex tasks like code generation and math problem-solving. Chain-of-Thought (CoT) reasoning~\cite{cot} has been crucial in breaking down problems into manageable steps, enhancing logical reasoning. Originally successful in AlphaGo's victory~\cite{alphazero}, MCTS has been adapted to guide model-based planning by balancing exploration and exploitation through tree-based search and random sampling, and later to large language model reasoning~\cite{sc-mcts, deepprover1.5, xu2023traingainunleashmathematical}.  However, recent research suggests that the vast token generation space of LLMs may make improving reasoning capabilities inefficient~\cite{deepseekai2025deepseekr1incentivizingreasoningcapability}. Additionally, long-path reasoning is likely influenced by the model's working memory~\cite{workingmemory}.

\paragraph{Large Language Model Post-training for Reasoning}
Recent work has focused on post-training strategies to enhance the reasoning capabilities of large language models, often through additional fine-tuning or reinforcement learning on specialized datasets with reasoning examples and chain-of-thought explanations~\cite{xu2025redstardoesscalinglongcot}. Reinforcement learning methods like Direct Preference Optimization (DPO)\cite{dpo}, Proximal Policy Optimization (PPO)\cite{ppo}, Group Relative Policy Optimization (GRPO)\cite{grpo}, and REINFORCE++\cite{rpp} are gaining attention. These strategies, alongside test-time scaling methods, are a promising frontier for advancing model reasoning.

\section{Comparion Between Base and Instruct Model}
We started with different model types (base, instruct) under the same training strategy and found that the RL curves of the two were surprisingly similar. The growth rates of the test set accuracy, curve, and reward curve were nearly the same. The slight difference is that the response length of the base model grows faster than that of the instruct model.
\label{base_comp}
\begin{figure}[H]
    \centering
    % 第一行，两张图片
    \begin{minipage}{0.48\textwidth}
        \centering
        \includegraphics[width=\textwidth, keepaspectratio]{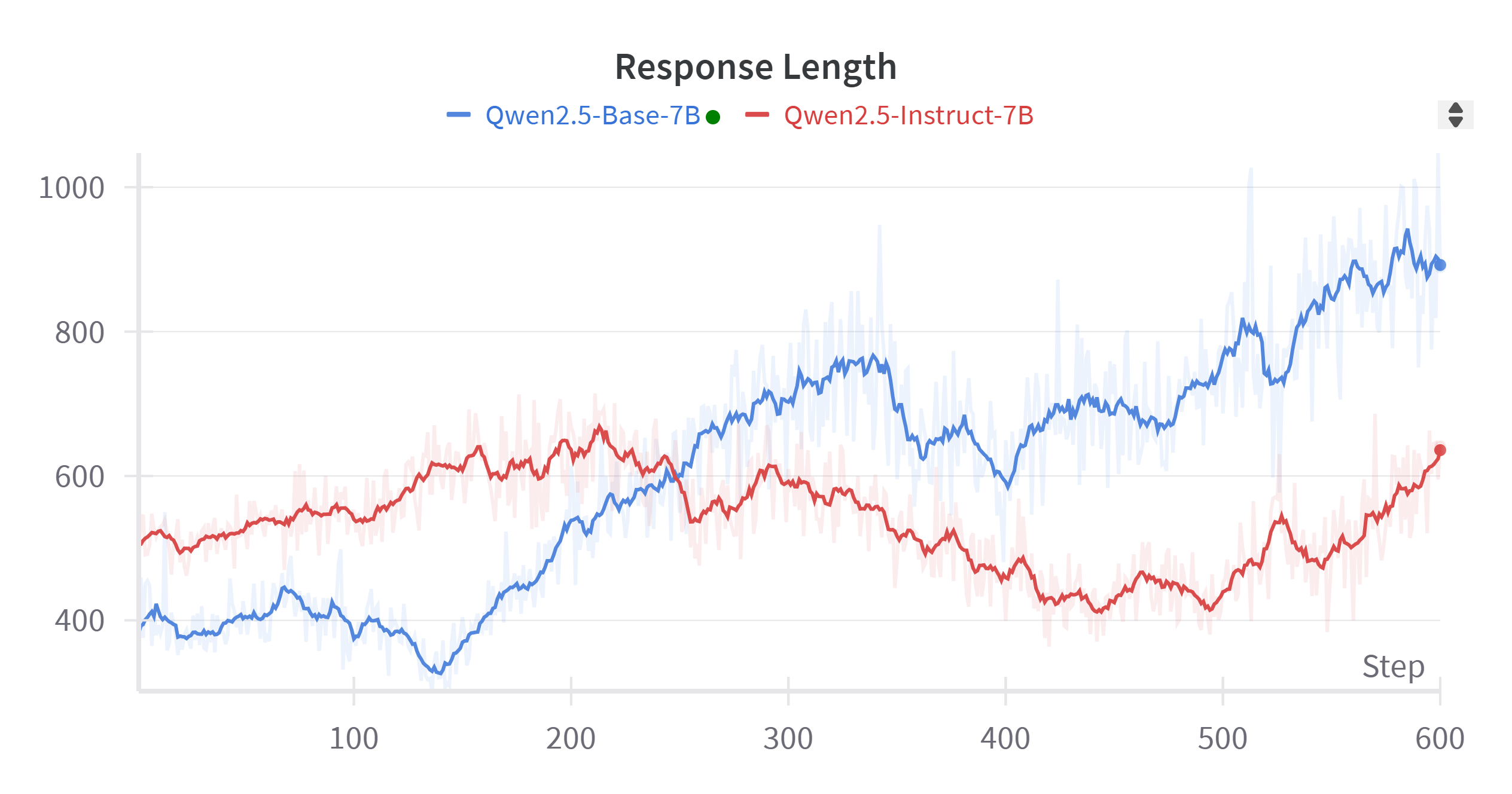}
    \end{minipage}
    \hfill
    \begin{minipage}{0.48\textwidth}
        \centering
        \includegraphics[width=\textwidth, keepaspectratio]{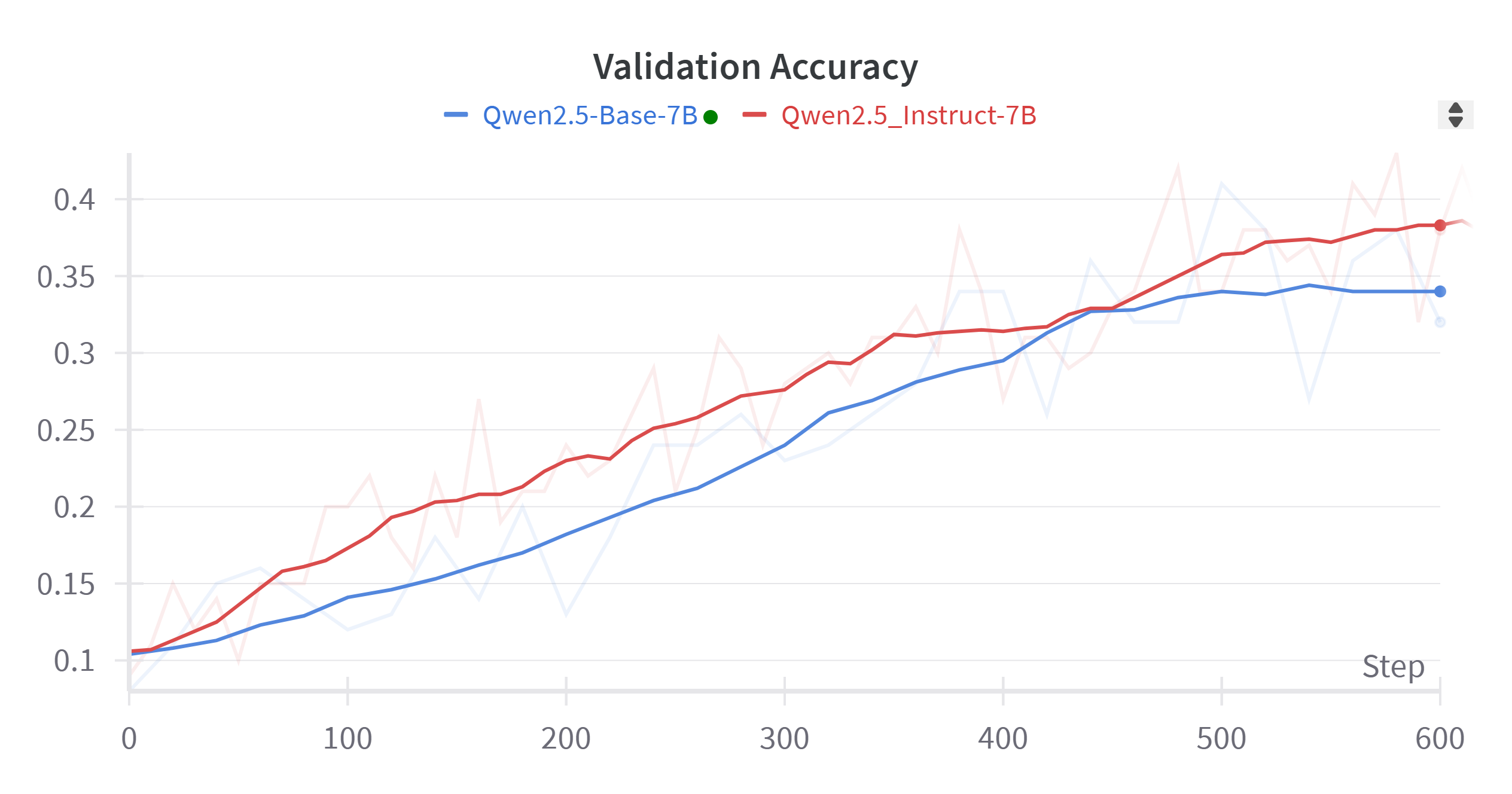}
    \end{minipage}

    % 第二行，两张图片
    \vspace{1em} % 添加一点垂直间距
    \begin{minipage}{0.48\textwidth}
        \centering
        \includegraphics[width=\textwidth, keepaspectratio]{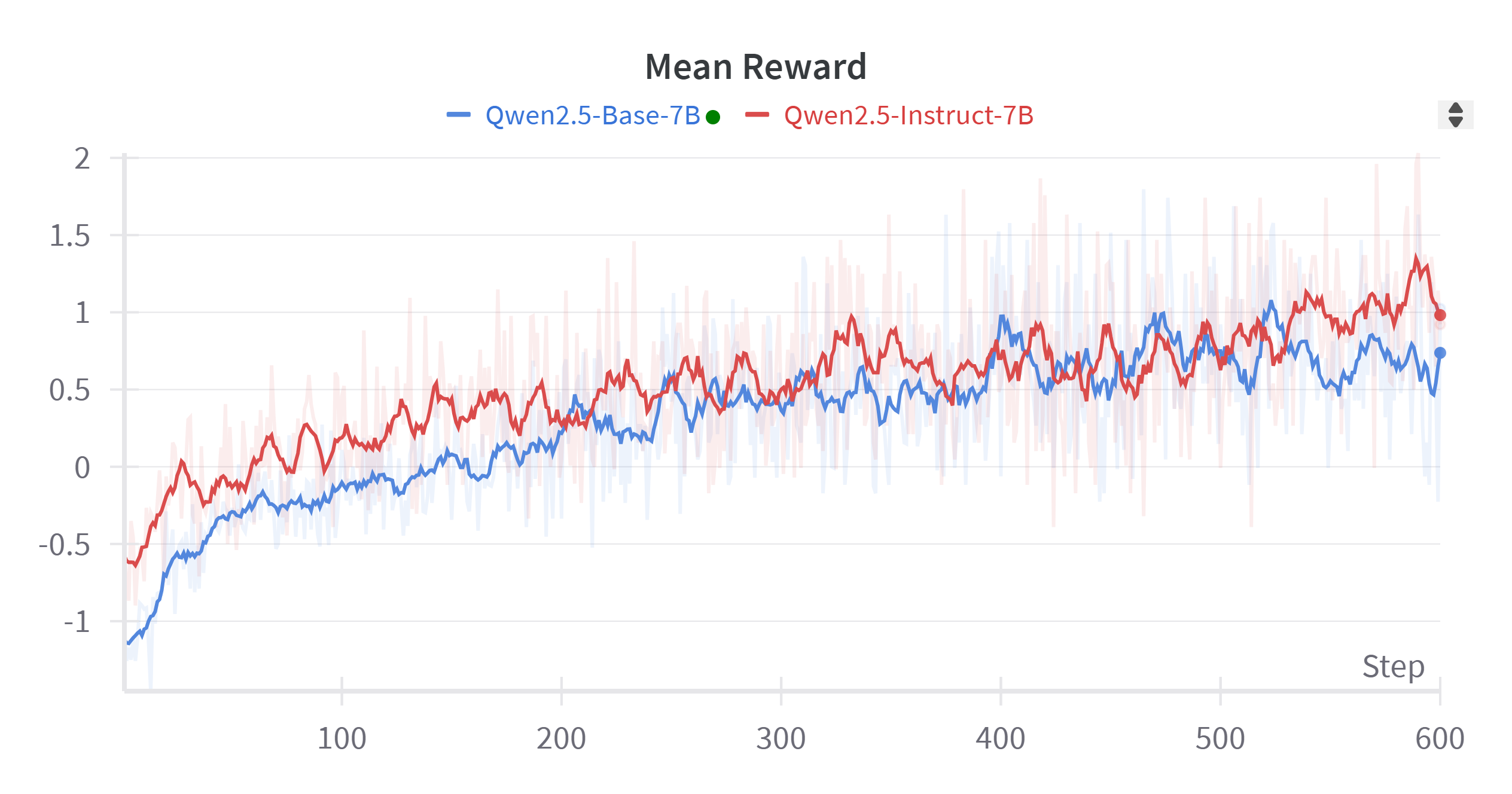}
    \end{minipage}
    \hfill
    \begin{minipage}{0.48\textwidth}
        \centering
        \includegraphics[width=\textwidth, keepaspectratio]{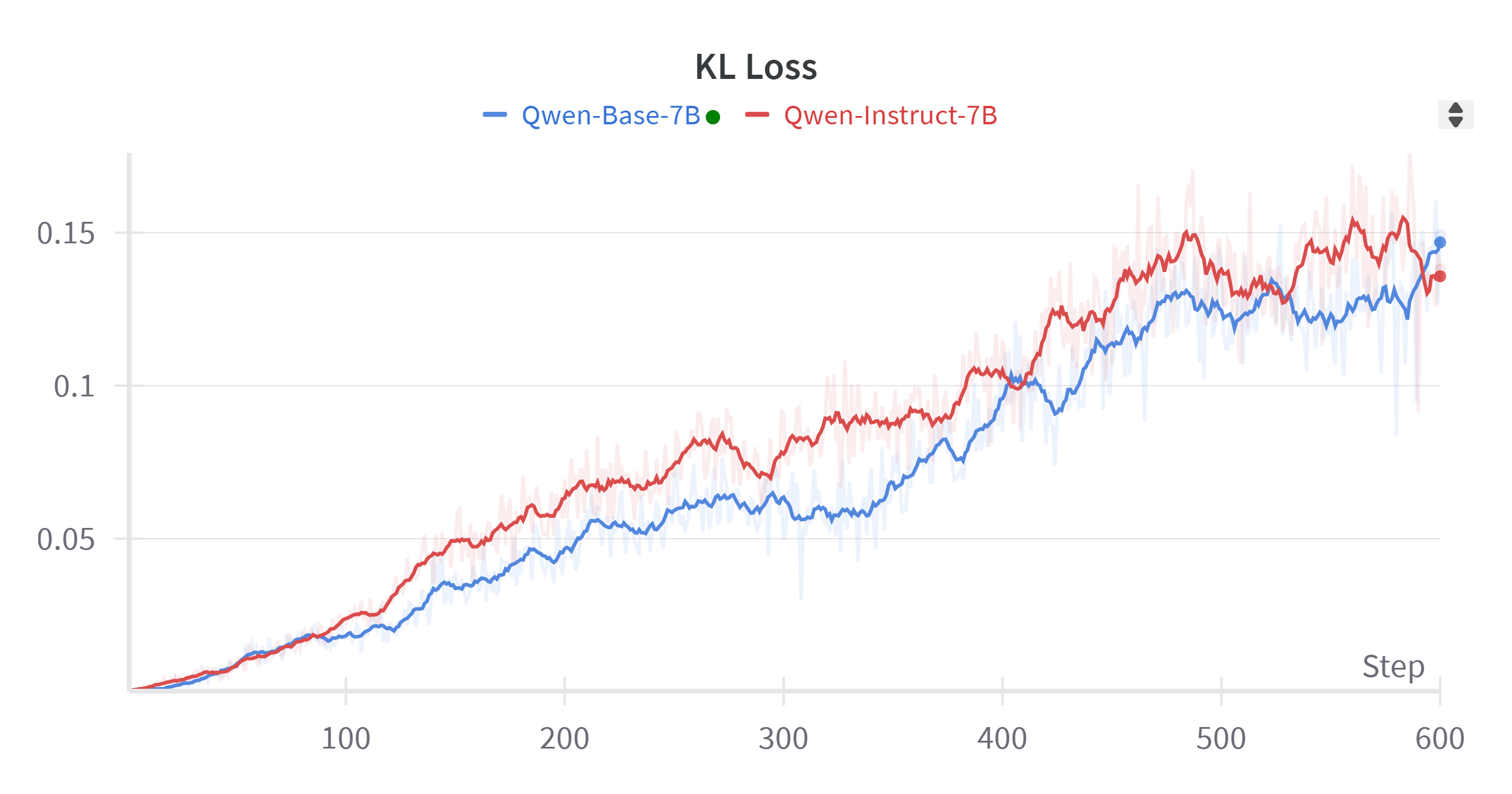}
    \end{minipage}

    % 添加一个大标题
    \caption{Comparison of training metrics: Response Length, Validation Accuracy, Mean Reward, and KL Loss.}
    \label{fig:training_metrics}
\end{figure}

\begin{figure}[H]
    \centering
    \includegraphics[width=\linewidth]{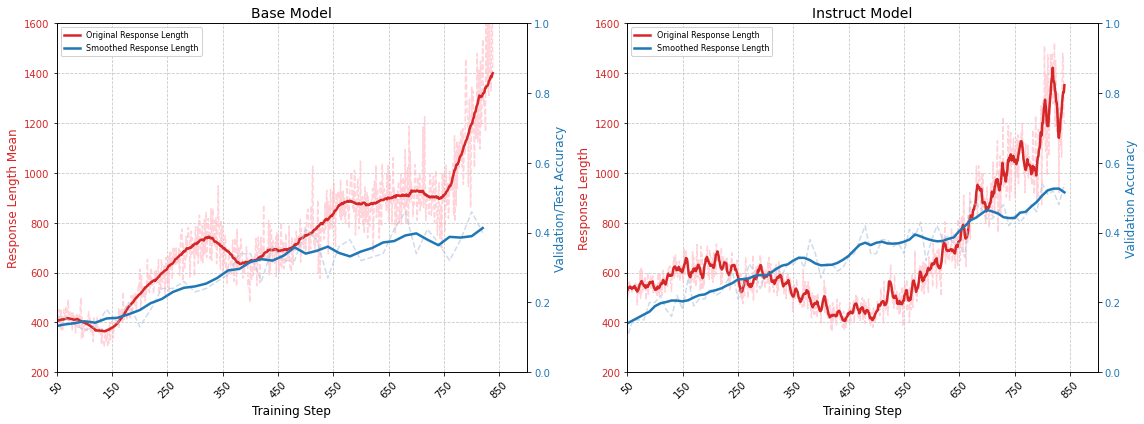}
    \caption{Length dynamics between base \& instruct}
    \label{fig:base_length}
\end{figure}
Figure \ref{fig:base_length} shows that the base model quickly increases thinking compute, while the instruct model consistently maintains slightly higher validation accuracy.

\section{Qualitative Analysis of Emergent Reasoning}
We highlight four key emergent behaviors observed in the RL-trained model:

\paragraph{(1) Hesitation and Self-Verification}
The model occasionally uses phrases like "I am not entirely sure; let's re-check this step" in the \texttt{<think>} section. This hesitation, absent in pre-training, emerges as the model is rewarded for correct answers and penalized for errors. Before providing the final \texttt{<answer>}, it systematically verifies all prior steps.

\paragraph{(2) Multi-Path Exploration \& Backtracking}
Encouraging the model to reason thoroughly, we see it propose multiple solutions (''Let's test both possibilitie'') and backtrack to check for consistency, resembling human problem-solving in logic puzzles.

\begin{table}[H]
\centering
\fontsize{9}{10}\selectfont
\begin{tabular}{p{14cm}}
\toprule
Question: \\
A very special island is inhabited only by knights and knaves. Knights always tell the truth, and knaves always lie. Five residents (William, Chloe, Lily, Jack, and Logan) made the following statements: (1) William said: "William is a knight and Chloe is a knight." (2) Chloe said: "Lily is a knight." (3) Lily said: "William is a knave and Jack is a knight." (4) Jack said: "William is a knave." (5) Logan said: "Lily is a knave if and only if William is a knave." So
who is a knight and who is a knave? \\
\midrule
Response: <think> \\
To solve this, let's start analyzing the statements one by one, beginning with William's statement since it's a good starting point due to its self-referential nature. \\
\\
William's statement: "William is a knight and Chloe is a knight." \\
   - If William is telling the truth (i.e., he is a knight), then both parts of his statement must be true. However, if William is a knight, the statement "William is a knight" is true, \textcolor{red}{but the statement "Chloe is a knight" cannot determine the truth about Chloe at this point since we haven't analyzed Chloe's statement yet.}\textcolor{blue}{\textit{\textbf{ (Hesitate)}}
} Therefore, William cannot be a knight because it would lead to a contradiction. \\
% - If William is lying (i.e., he is a knave), then at least one part of his statement is false. \\
\ldots \\

% Since "William is a knight" is false, this part is indeed false, so his statement is technically true in the context of him being a knave, which is a contradiction. Therefore, William must be a knave. \\
% \\
% \textbf{Now that we know William is a knave, let's analyze Chloe's statement}: "Lily is a knight." \\
%    - Since William is a knave, Chloe's statement can be either true or false. If Chloe is a knight, her statement must be true, meaning Lily is a knight. If Chloe is a knave, her statement must be false, meaning Lily is not a knight. \\
   
\textcolor{red}{Let's test both possibilities.}\textcolor{blue}{\textit{\textbf{ (Explore)}}}\\
Testing Chloe as a knight: If Chloe is a knight, then her statement "Lily is a knight" is true, so Lily is a knight. \\
Analyzing Lily's statement: "William is a knave and Jack is a knight." \\
\ldots \\

%    - Since William is a knave and Lily is a knight, her statement is true. This means Lily is indeed telling the truth, so she is a knight. \\
% \\
% \textbf{Analyzing Jack's statement}: "William is a knave." \\
%    - Since William is a knave, Jack's statement is true, so Jack is a knight. \\
% \\
% \textbf{Analyzing Logan's statement}: "Lily is a knave if and only if William is a knave." \\
%    - Since Lily is a knight and William is a knave, this statement is false. Therefore, Logan is a knave. \\
% \\
% \textbf{Analyzing Chloe's statement again}: We already determined that if Chloe is a knight, her statement "Lily is a knight" is true, so she is a knight. \\
%    - Since all statements are consistent with Chloe being a knight, Chloe is indeed a knight. \\
% \\
\textcolor{red}{Now, we have determined:}\textcolor{blue}{\textit{\textbf{ (Summerize)}}} \\
- William is a knave. \\
- Chloe is a knight. \\
\ldots \\
\textcolor{red}{Let's verify all statements:} \textcolor{blue}{\textit{\textbf{ (Verify)}}} \\
- "William is a knight and Chloe is a knight" (False, since William is a knave). \\
- "Lily is a knight" (True). \\
\ldots \\
% - "Lily is a knave if and only if William is a knave" (False, since Lily is a knight and William is a knave). \\
% - "William is a knave" (True). \\
% - "William is a knave and Jack is a knight" (True). \\
% \\
% All statements are consistent with our assignments. \textless\textbackslash think>
\\
\hline
\end{tabular}
\vspace{5mm}
\caption{An example of a large language model trained through reinforcement learning naturally exhibiting reasoning behaviors such as hesitation, exploration, summarization, and verification, even though these behaviors were not present in the training data.}
\end{table}

\paragraph{(3) Formula Application}
After RL training, our model instinctively applied the "If P, then Q" implication formula when solving logical puzzles, like the Knights and Knaves problem. This formula asserts that the proposition is false only when P is true and Q is false. We were surprised to see that the model not only solved the puzzles through trial and error but also incorporated formal logical reasoning, resembling human problem-solving, despite no such data included in the training set.

\begin{table}[H]
\centering
\fontsize{9.5}{10}\selectfont
\begin{tabular}{p{14cm}}
\toprule
Question: \\
A very special island is inhabited only by knights and knaves. Knights always tell the truth, and knaves always lie. Three residents (Jack, Samuel, Lily) made the following statements: (1) Jack said, "If Samuel is a knight, then Lily is a knave." (2) Samuel said, "If Jack is a knight, then Lily is a knave." (3) Lily said, "Jack is a knight." So who is a knight and who is a knave? \\
\midrule
Response: <think> \\
Let's analyze the statements step by step. \\
1. Jack's statement is: "If Samuel is a knight then Lily is a knave." \\
\ldots \\
Now that we know Samuel is a knave, let's analyze Samuel's statement again: \\
- Samuel's statement: "If Jack is a knight then Lily is a knave." \\
\textcolor{red}{- An implication "If P then Q" is false only when P is true and Q is false.}\textcolor{blue}{\textit{\textbf{ (Formula Application)}}} Here, "If Jack is a knight then Lily is a knave" is false when Jack is a knight and Lily is not a knave. This is consistent with our previous conclusion that Jack is a knight and Lily is a knight. \\
\ldots \\
\hline
\end{tabular}
\vspace{5mm}
\caption{An example of using logical formulas to solve a Knights and Knaves puzzle.}
\end{table}

\paragraph{(4) Occasional Language Switching}
Interestingly, some \texttt{<think>} segments contain Chinese tokens (the base model is English-centric). The final \texttt{<answer>} remains in English, presumably to gain a format reward. In some demos, the model briefly interjects a line of Chinese while analyzing statements, then seamlessly switches back to English for the solution. This phenomenon may indicate that the model uses language mixing as a hidden fallback or distinct internal representation. 

\begin{table}[H]
\centering
\fontsize{9.5}{10}\selectfont
\begin{tabular}{p{14cm}}
\toprule 
Question:A very special island is inhabited only by knights and knaves. Knights always tell the truth, and knaves
always lie. Six residents (Benjamin, Michael, Evelyn, Sophia, James, and Jacob) made the following statements: (1) Benjamin said, "If Sophia is a knight, then Evelyn is also a knight." (2) Michael said, "Benjamin is a knave, or James is a knight." (3) Evelyn said, "Sophia is a knight, or James is a knave." (4) Sophia said, "If Michael is a knave, then James is a knight." (5) James said, "Sophia is a knight, or Evelyn is a knight." (6) Jacob said, "Benjamin is a knave, and Sophia is a knight." So who is a knight and who is a knave? \\
\midrule
Response: <think> \\
Let's analyze each statement step-by-step based on whether the speaker is a knight or a knave. \\

1. Benjamin: "If Sophia is a knight then Evelyn is a knight" \\
   - If Benjamin is a knight, then the statement must be true. Thus, if Sophia is a knight, Evelyn has to be a knight. \\

2. Michael: "Benjamin is a knave or James is a knight" \\
\ldots \\

From Jacob's statement, if Jacob is a knight then Sophia must be a knight and Benjamin must be a knave. \\

\begin{CJK*}{UTF8}{gbsn}
\fontsize{8.5}{10}\selectfont
\textcolor{red}{以卡文的方式推理: }

\textcolor{red}{- 假设 Jacob 是骑士，那么 Jacob 的声明 "Benjamin 是背叛者且 Sophia 是骑士" 就是真的。这意味着 Benjamin 是背叛者而 Sophia 是骑士。}

\textcolor{red}{- 既然 Sophia 是骑士，根据 Benjamin 的声明 "如果 Sophia 是骑士那么 Evelyn 也是骑士"，Evelyn 也是骑士。}

\ldots

\textcolor{red}{综上所述，Sophia, Evelyn, 和 Jacob 是骑士，而 Benjamin, Michael, 和 James 是背叛者。 }\textcolor{blue}{\textit{\textbf{ (Language Switching)}}}

% \textless\textbackslash think>
\end{CJK*}
\ldots \\
\hline
\end{tabular}
\vspace{5mm}
\caption{An example of our model used language switching to solve a Knights and Knaves puzzle.}
\end{table}

\end{document}